\documentclass[10pt,twocolumn,letterpaper]{article}

\usepackage{cvpr}              

\usepackage{xspace}

\usepackage{colortbl}
\definecolor{bestcolor}{gray}{.9}
\newcommand{\bestcell}[1]{\cellcolor{bestcolor}{#1}}

\usepackage{changepage}

\usepackage{amsmath}
\usepackage{amssymb}
\usepackage{mathtools}
\usepackage{pifont}
\usepackage{amsthm}
\usepackage{makecell}
\usepackage{multirow}
\usepackage{tcolorbox}          
\usepackage{listings}           
\tcbuselibrary{listings}        
\tcbuselibrary{breakable}       

\newtcblisting{promptcode}[1][Python Prompt Example]{
    listing only,
    listing options={
        breaklines=true,       
        breakatwhitespace=true,
        keepspaces=true,       
        basicstyle=\ttfamily\scriptsize, 
        showstringspaces=false,
        tabsize=4,            
        columns=flexible,      
        escapeinside={/*}{*/},  
    },
    breakable=true,           
    width=\linewidth,         
}

\newtcblisting{pythoncode}[1][Python Prompt Example]{
    listing only,
    listing options={
        language=Python,
        breaklines=true,       
        breakatwhitespace=true,
        keepspaces=true,       
        basicstyle=\ttfamily\scriptsize, 
        showstringspaces=false,
        tabsize=4,            
        columns=flexible,      
        escapeinside={/*}{*/},  
    },
    breakable=true,           
    width=\linewidth,         
}

\theoremstyle{plain}

\theoremstyle{definition}

\theoremstyle{remark}

\usepackage{algorithm}
\usepackage{algpseudocode}
\algrenewcommand\algorithmicrequire{\textbf{Input:}}
\algrenewcommand\algorithmicensure{\textbf{Output:}}

\definecolor{cvprblue}{rgb}{0.21,0.49,0.74}

\usepackage[pagebackref,breaklinks,colorlinks,allcolors=cvprblue]{hyperref}
\hypersetup{linkcolor={red}}  
\usepackage{graphicx}
\usepackage{caption}
\usepackage{float}
\def\paperID{9694} 
\def\confName{CVPR}
\def\confYear{2025}

\title{
RoboTwin: Dual-Arm Robot Benchmark with Generative Digital Twins
}

\author{%
  Yao Mu$^{1,3*\dag}$\quad Tianxing Chen$^{1,3,4*}$\quad Zanxin Chen$^{2,4*}$\quad Shijia Peng$^{2,4*}$\quad {Zhiqian Lan}$^1$\\ {Zeyu Gao}$^{5}$\quad {Zhixuan Liang}$^1$\quad
  {Qiaojun Yu}$^9$\quad {Yude Zou}$^{4}$\quad {Mingkun Xu}$^7$\\ {Lunkai Lin}$^{2}$\quad {Zhiqiang Xie}$^{2}$\quad {Mingyu Ding}$^{6}$\quad {Ping Luo}$^{1,8\dag}$ \\
  [1.5mm]
$^1$HKU \quad $^2$Agilex Robotics \quad $^3$Shanghai AI Laboratory \quad
  $^4$SZU \\ $^5$CASIA \quad $^6$UNC-Chapel Hill \quad $^7$GDIIST \quad
 $^8$HKU-Shanghai ICRC \quad $^9$SJTU\\
  \normalsize{\url{https://robotwin-benchmark.github.io}}
}

\begin{document}
\maketitle

\footnotetext{$^*$~Equal contribution. $\dag$~Corresponding authors.}~

\begin{abstract}
In the rapidly advancing field of robotics, dual-arm coordination and complex object manipulation are essential capabilities for developing advanced autonomous systems. 
%
However, the scarcity of diverse, high-quality demonstration data and real-world-aligned evaluation benchmarks severely limits such development.
To address this, we introduce RoboTwin, a generative digital twin framework that uses 3D generative foundation models and large language models to produce diverse expert datasets and provide a real-world-aligned evaluation platform for dual-arm robotic tasks.
Specifically, RoboTwin creates varied digital twins of objects from single 2D images, generating realistic and interactive scenarios. It also introduces a spatial relation-aware code generation framework that combines object annotations with large language models to break down tasks, determine spatial constraints, and generate precise robotic movement code.
Our framework offers a comprehensive benchmark with both simulated and real-world data, enabling standardized evaluation and better alignment between simulated training and real-world performance.
We validated our approach using the open-source COBOT Magic Robot platform. Policies pre-trained on RoboTwin-generated data and fine-tuned with limited real-world samples demonstrate significant potential for enhancing dual-arm robotic manipulation systems by improving success rates by over 70\% for single-arm tasks and over 40\% for dual-arm tasks compared to models trained solely on real-world data.
%


\end{abstract}    
\section{Introduction}

%
Robotic systems with intricate dual-arm coordination and precise dexterity are essential for complex object manipulation to unlock advanced capabilities across domains such as healthcare, manufacturing, logistics, and domestic assistance.
However, creating robust and versatile robotic systems that meet these demands remains a challenge, with a major bottleneck being the absence of diverse, high-quality training data and comprehensive evaluation benchmarks that are aligned with the real world.


\begin{figure}[t] 
    \centering
    \includegraphics[width=0.96\linewidth]{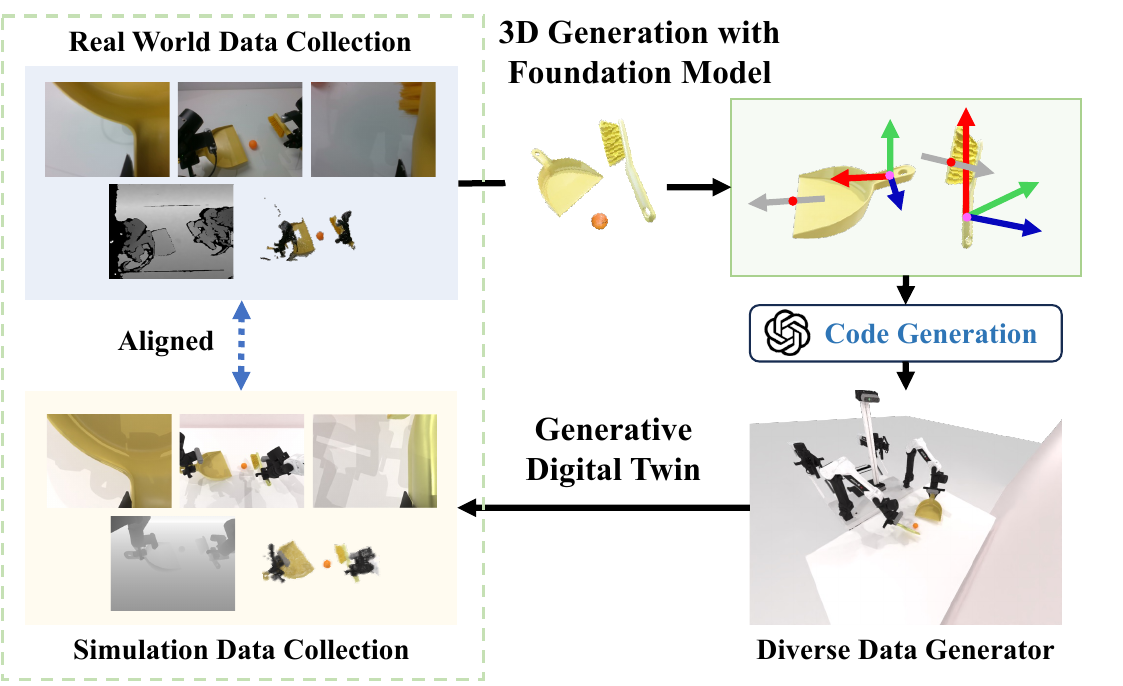}
    \vspace{-8pt}
    \caption{\textbf{RoboTwin Benchmark.} A framework leveraging generative foundational models to generate realistic and interactive training scenarios and diverse expert demonstrations for benchmarking dual-arm robotic manipulation.}
     \vspace{-16pt}
    \label{fig:RoboTwin Benchmark} 
\end{figure}

Traditional approaches to data collection, particularly human teleoperation~\cite{ebert2021bridge, jang2022bcz,fu2024mobile,aldaco2024aloha,ding2024bunny,cheng2024open}, yield high-quality demonstrations but face significant practical limitations.
While these methods provide reliable training data, they are often prohibitively expensive, time-intensive, and struggle to cover the diverse range of scenarios robots encounter in real-world deployments. 
To address these limitations, researchers have turned to algorithmic trajectory generators in simulations~\cite{jiang2023vima, gu2023maniskill2, dalal2023imitating}. These alternatives, however, frequently require task-specific design, hindering their generalizability and scalability. 
Recent advances such as MimicGen~\cite{mandlekar2023mimicgen} and RoboCaca~\cite{nasiriany2024robocasa} have demonstrated significant progress in generating large-scale simulated expert data from limited human demonstrations. However, these approaches operate under fixed scenario settings and struggle to handle task variants beyond their predefined configurations, limiting their generalizability to novel scenarios.

Another limitation of existing benchmarks is that they predominantly focus on single-arm tasks ~\cite{gu2023maniskill2,mees2022calvin} or bimanual tasks with two separated arms~\cite{grotz2024peract2}, which fail to capture the complexity and coordination requirements inherent in integrated dual-arm systems. 
While HumanoidBench~\cite{sferrazza2024humanoidbench} and BiGym~\cite{chernyadev2024bigym} explore benchmarks for humanoid bimanual manipulation, their scalability is limited by fixed environments or reliance on VR teleoperation for demonstration collection.
As a result, these gaps highlight the urgent need for a scalable and standardized dual-arm collaboration benchmark with an efficient data collection pipeline.

%


To address these challenges, as shown in Fig. \ref{fig:RoboTwin Benchmark}, we propose RoboTwin, a generative digital twin framework empowered by 3D generative foundation models and large language models (LLMs), aiming to produce diverse expert datasets and provide a real-world-aligned evaluation platform for dual-arm robotic tasks.
Starting from a single 2D RGB image, we employ generative foundation models for 3D modeling and texture generation, enabling the efficient creation of varied object instances with different shapes, sizes, and appearances.
Each object class is incorporated with spatial annotations, which define function axes, approach axes, lateral axes, and contact points and are applicable across various instances within an object class via feature point matching technology.
Building upon these spatially-aware digital twins, RoboTwin leverages LLMs to interpret and decompose complex tasks into manageable sub-tasks. For each sub-task, we infer the constraints of the terminal state. For example, in a hammering task, the functional point of the hammer head needs to align with the surface of the target object.
RoboTwin then generates executable code that calculates key poses based on these spatial constraints and object properties, interfacing with underlying planning modules to produce complete, feasible trajectories for execution.

Within the above framework, our RoboTwin features diverse dual-arm manipulation tasks that combine simulated expert data with real-world teleoperated datasets under consistent environmental and hardware setups.
%
%
%
We then benchmark and evaluate the ability of RoboTwin to improve policy generalization in real-world scenarios. 
Experimental results demonstrated that policies pre-trained on 300 RoboTwin-generated samples and fine-tuned with 20 real-world samples improve the success rate by 70\% in single-arm manipulation tasks like hammer beat, and over 40\% in dual-arm coordination tasks, such as ball sweep, compared to those trained exclusively on 20 real-world samples.
%

We summarize our key contributions as: 
1) we establish a convenient real-to-sim pipeline that requires only an RGB image from the real world to generate diverse 3D models of target objects, empowered by a 3D generative foundation model; 
2) we create a spatial-aware code generation framework, which automatically creates expert-level demonstration data via a large language model and the spatial annotations of the target objects.
3) we develop a standard benchmark for dual-arm manipulation tasks including both real-world teleoperated data and high-fidelity synthetic data generated for corresponding scenarios.
These advancements provide a robust framework for generating diverse, high-quality training data and policy evaluation for dual-arm manipulation tasks, significantly contributing to the development of more capable and versatile robotic systems.


\begin{figure*}[t]
    \centering 
    \includegraphics[width=0.9\textwidth]{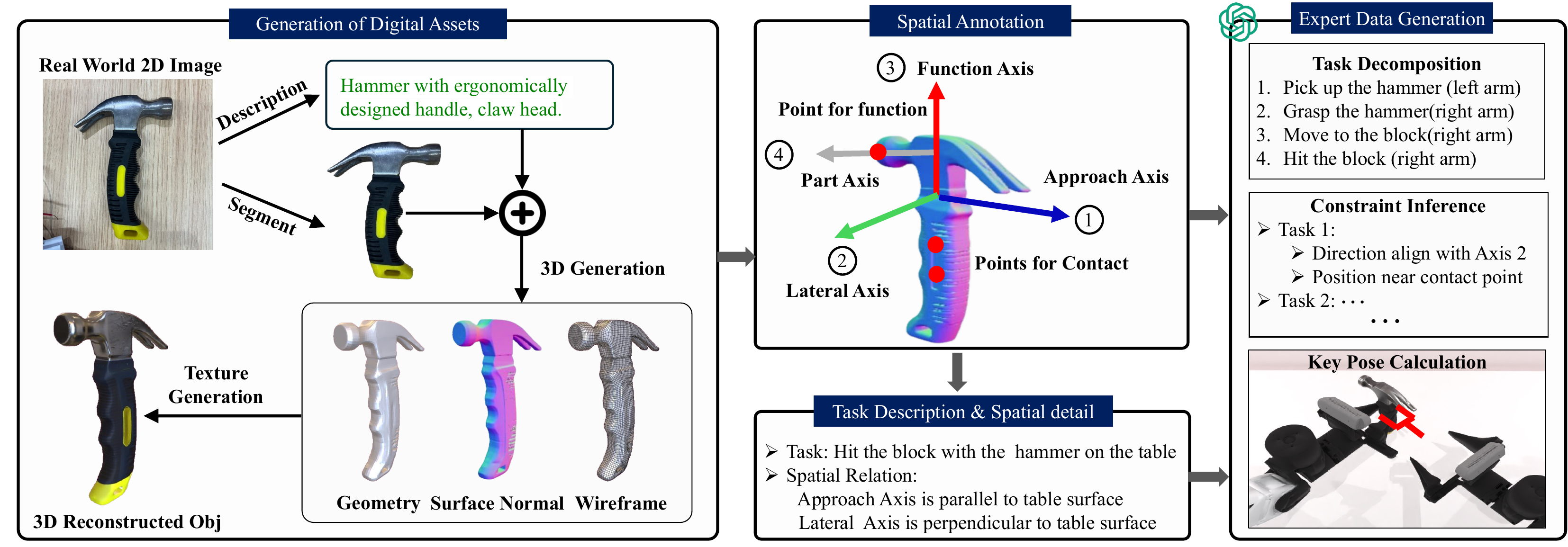}
    \vspace{-7pt}
\caption{\textbf{Real-to-simulation transfer and expert data generation.} We first leverage a 3D generative foundation model to create diverse 3D assets from 2D images, complete with geometry, normals, and textures. This process is augmented by vision-language models to generate variations of object descriptions, enabling the creation of visually diverse yet functionally consistent 3D models. We then implement a spatial annotation framework that marks key functional and contact points, along with functional, approach, and lateral axes on these 3D assets. Finally, we employ LLMs to generate expert demonstrations by decomposing tasks into subtasks, inferring spatial constraints, and generating collision-free robot behavior executable code that satisfies kinematic requirements.}
    \label{fig:aigc} 
    \vspace{-8pt}
\end{figure*}

\section{Related Work}

\subsection{Datasets and Benchmarks for Robotics}
To collect effective demonstrations for robotic tasks, human teleoperation is the most common approach, where human manually guides a robot across various tasks~\cite{mu2024robotwin, mandlekar2018roboturk, mandlekar2019scaling, mandlekar2020hitl, ebert2021bridge, jang2022bcz}. Recent advancements have extended this methodology by employing teams of human operators over prolonged periods to assemble substantial real-world datasets~\cite{ebert2021bridge, ahn2022can, jang2022bcz, rt12022arxiv}.
%
An alternative method employs algorithmic trajectory generators within simulators~\cite{james2020rlbench, zeng2020transporter, jiang2023vima, gu2023maniskill2, dalal2023imitating}. Nevertheless, such approaches typically demand manual, task-specific design for individual tasks.
Recent initiatives like MimicGen~\cite{mandlekar2023mimicgen} and RoboCaca~\cite{nasiriany2024robocasa} generate simulated expert data by adapting actions to new object poses, but remain limited to fixed scenarios and predefined task configurations.
Furthermore, their reliance on fixed 3D objects limits the diversity of interacting objects and shapes.
Besides, Maniskill~\cite{gu2023maniskill2,taomaniskill3} provides diverse simulation scenarios but lakes automated data collection mechanism.


In contrast, RoboTwin leverages 3D generative foundation models and LLMs to autonomously create both task variations and corresponding expert demonstrations. 
From 3D assets, it generates task scenarios and executable code via spatial reasoning, minimizing human intervention and supporting diverse object appearances.

\subsection{Dual-arm Manipulation}

While significant advances have been made in single-arm manipulation, coordinated multi-arm manipulation remains largely unexplored. Peract2~\cite{grotz2024peract2} offers benchmarks for bimanual tasks with separated arms, but its setup lacks the complexity of integrated dual-arm systems. HumanoidBench~\cite{sferrazza2024humanoidbench} evaluates dexterous, whole-body manipulation with a humanoid robot in a fixed reinforcement learning benchmark, while BiGym~\cite{chernyadev2024bigym} provides a bimanual benchmark but is constrained by VR teleoperation, limiting their scalability in data collection and evaluation.
As a benchmark for dual-arm tasks, RoboTwin enables automatic and large-scale coordinated manipulation data generation with comprehensive policy evaluation.

\subsection{Robot Manipulation Learning Methods}
The adoption of human demonstrations to instruct robots in manipulation skills is a prevalent method in Robot Manipulation Learning \cite{sharma2018multiple,bahl2023affordances,jang2022bc,lynch2023interactive,brohan2022rt,chi2023diffusion,ze20243d,chen2024g3flow,lu2024manicmrealtime3ddiffusion,fu2025cordvipcorrespondencebasedvisuomotorpolicy}. Among the techniques, Behavioral Cloning stands out for learning policies offline from these demonstrations. It replicates observed actions from a curated dataset \cite{pomerleau1989alvinn,zhang2017deep, mandlekar2020learning, ebert2021bridge, rt12022arxiv, jang2022bcz, dalal2023imitating, jiang2023vima}. Conversely, Offline Reinforcement Learning enhances policy learning by optimizing actions based on a predefined reward function and exploiting large datasets \cite{levine2020offline,kalashnikov2021mt, chebotar2023q, gurtler2023benchmarking, kumar2022pre, kumar2021workflow, liu2025avractivevisiondrivenrobotic}. The Action Chunking with Transformers (ACT) technique integrates a Transformer-based visuomotor policy with a conditional variational autoencoder to structure the learning of action sequences ~\cite{zhao2023learning,vaswani2017transformer,sohn2015learning}. Diffusion models have been introduced into robot imitation learning and are gradually becoming a mainstream approach due to their excellent generative capabilities~\cite{janner2022planning, decisiondiffuser, adaptdiffuser,metadiffuser, skilldiffuser,liang2024dexdiffuser}. Recently, the Diffusion Policy method has gained prominence. It employs a conditional denoising diffusion process for visuomotor policy representation, effectively reducing the accumulative error in trajectory generation that is often observed in Transformer-based visuomotor policies \cite{chi2023diffusion}. The 3D Diffusion Policy \cite{ze20243d} uses point clouds for environmental observations, enhancing spatial information utilization and managing various robotic tasks in both simulated and real environments with only a small number of demonstrations.



\subsection{LLM for Robotic Code Generation.}
With their remarkable ability in natural language understanding and code generation, Large Language Models (LLMs) have revolutionized numerous domains in artificial intelligence. In robotics, these models have shown exceptional capabilities in bridging the gap between natural language commands and executable robot actions~\cite{driess2023palm,huang2023voxposer,hu2023tree,huang2024rekep,mu2024embodiedgpt,hu2023look,huang2024copa,liu2024moka,chen2024roboscript,gao2024dag,chen2024vlmimic,sha2023languagempc,wu2024plot2code,liangdexhanddiff}. Code as Policies~\cite{liang2023code} and RoboCodeX~\cite{robocodex,chen2024roboscript} established that LLMs can effectively translate high-level task descriptions into functional robot control programs. While Rekep~\cite{huang2024rekep} advances spatial reasoning between key points, it has limitations in handling functional axis constraints and fails to account for spatial relationships between object functional axes and the table surface during code generation. Furthermore, existing code generation approaches predominantly focus on single-arm robots, overlooking crucial aspects of dual-arm collaboration and active collision avoidance strategies.
\section{Bridging Physical and Digital Worlds for Diverse Robot Behavior Generation}
\label{section:method}
\begin{figure*}[t]
    \centering 
    \includegraphics[width=0.95\textwidth]{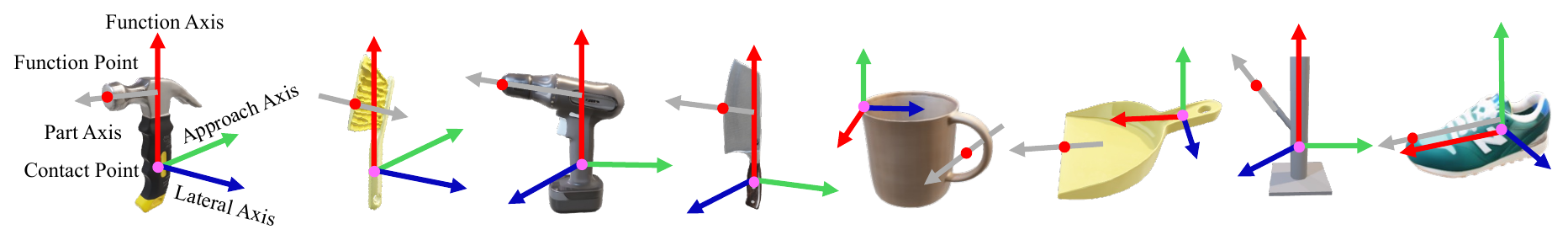}
    \vspace{-12pt}
    \caption{\textbf{Examples of spatial annotations.} Function and contact points with principal axes for functional parts and approach directions are extracted semi-automatically within RoboTwin for spatial- and geometry-aware manipulation and code generation.}
    \label{fig:Spatial Annotation Framework} 
    \vspace{-12pt}
\end{figure*}

\subsection{Generation of Diverse Digital Assets}

\footnotetext[4]{We use Deemos's 3D digital asset Generation Model (from text or image) Rodin: \href{https://hyperhuman.deemos.com/rodin}{https://hyperhuman.deemos.com/rodin}}

Our approach utilizes Deemos's Rodin platform\footnotemark[4] to create 3D models from simple 2D RGB images. This method significantly reduces the need for expensive sensors while achieving realistic visual effects and supporting physical simulations. The process begins with capturing photographs of real-world objects. 
As shown in Fig.~\ref{fig:aigc}, we use GPT-4V~\cite{2023GPT4VisionSC} to analyze these images to generate corresponding descriptions, which are then autonomously modified via language model to create similar yet visually distinct object descriptions. We use these descriptions with SDXL-Turbo~\cite{SDXL} to generate a diverse set of 2D images representing various appearances of the same object class. An image-conditioned 3D generation model then processes this collection of images, producing a wide range of 3D models for a single object type. The final output transforms a 2D image into a comprehensive 3D model, featuring detailed geometry, surface normals, wireframes, and textures. We validate asset quality using two complementary approaches: quantitative evaluation via UCLIP-I~\cite{clay} similarity metrics and qualitative assessment through GPT-4V visual validation. Assets falling below quality thresholds are automatically flagged for regeneration. This dual validation approach ensures both visual and geometry consistency for effective sim-to-real transfer. To ensure physical fidelity, our pipeline leverages GPT-4V to classify object materials and assign appropriate physics parameters with $\pm5\%$ random variations to enhance robustness. 

\subsection{Spatial Annotation Framework for 3D Assets}

To enhance the structural integrity and universal applicability of generated assets, we implement a systematic approach for annotating key points and axes on tools. This methodology aims to render the data more comprehensible and accessible to large language models for complex task code generation. As shown in Fig.~\ref{fig:Spatial Annotation Framework}, the annotation process focuses on two primary elements: key points and axes. 

\textbf{Key Points.}
Key points represent specific locations on tools directly associated with their functional operations or user interaction points. We distinguish between these two types:
\textbf{(1) Point for Function}: This key point designates the primary functional component of the tool, such as the striking surface of a hammer. It defines the tool's functional origin or point of action, directly correlating to the tool's primary purpose in a given task.
\textbf{(2) Point for Contact}: This key point indicates the area of interaction between the tool and its user or other objects. It represents the gripping point or contact area, serving as a crucial human-machine interface point. Annotating this point facilitates understanding of tool's operational posture.

\textbf{Axes.}
Axes are used to describe the spatial directionality of tools during task execution, encompassing the direction of functional execution and the tool's approach towards objects. We identify three principal axes:
\textbf{(1) Function Axis}: This axis represents the direction in which the tool executes its primary function. It typically aligns with the tool's main operational vector, guiding the understanding of the tool's intended use and movement during task performance.
\textbf{(2) Approach Axis}: The approach axis delineates the direction in which the tool approaches or is applied to the target object. This axis is crucial for comprehending the spatial relationship between the tool and its subject of operation.
\textbf{(3) Lateral Axis}: This axis is perpendicular to both the function and approach axes, completing a three-dimensional coordinate system for the tool. The lateral axis aids in defining the tool's orientation and potential rotational movements during use.

By systematically annotating these key points and axes, we create a comprehensive spatial framework for each tool. This framework enables a more precise and context-aware understanding of tool functionalities, facilitating improved task planning and execution by large language models. We do not need to repeatedly annotate different 3D models from the same class. Instead, to streamline the annotation process for various 3D models of similar objects, we employ a feature point matching approach leveraging the Stable Diffusion~\cite{rombach2022high} encoder. This method enables the transfer of key points across various 3D models within the same object class. Our approach utilizes feature point matching to determine the target point. 
Specifically, under the table top view, given a source image $I_s$, a target image $I_t$, and a source point $p_s$, we aim to locate the corresponding point $p_t$ in the target image.
Following the methodology outlined in ~\cite{tang2023emergent,ju2024robo}, we extract diffusion features from both $I_s$ and $I_t$. Since these diffusion features correspond to individual pixels in the target image, we can identify the pixel in $I_t$ with the highest similarity to $p_s$ by analyzing the extracted features. 
This technique allows for efficient key point migration across different 3D models of similar objects, eliminating the need for redundant annotations and enhancing the overall efficiency of the 3D modeling process. 

\subsection{Expert Data Generation}
\label{sec:code generation}
Building upon our spatial annotation framework and expert data generation pipeline, we present a systematic approach to generating robot behaviors that satisfy spatial constraints while ensuring collision-free execution. At the core of our framework lies a comprehensive dual-arm manipulation system with three key capabilities. First, it enables synchronized arm movements through screw motion interpolation coupled with coordinated gripper actions, ensuring stable object handling. Second, it supports independent arm operations for scenarios requiring asymmetric movements. Third, it implements dynamic collision avoidance through continuous adjustment of safe intermediate positions between arms. Our motion generation implements a three-stage approach: (1) spatial constraint inference that analyzes object annotations to establish geometric relationships, (2) LLM-based code generation translating constraints into executable code using the MPlib trajectory optimization library, and (3) execution validation ensuring task completion. We incorporate a self-correction mechanism where execution errors are fed back to the language model, with minimal human oversight for complex cases. 
Leveraging these integrated capabilities, we employ large language models (LLMs) with predefined APIs to systematically generate expert demonstrations across diverse robotic tasks. The process consists of the following detailed steps:


\begin{enumerate}
    \item \textbf{Scene Initialization}: The task environment is set up with relevant objects and their initial poses. For instance, a hammering task would involve placing the hammer and target objects in their starting positions.
    
    \item \textbf{Task Decomposition}: Based on human input describing the task, we use LLM to break it down into subtasks. For example, a ``hammer a nail" task might be decomposed into: a) grasping the hammer, b) positioning the hammer over the nail, c) striking the nail, and d) returning the hammer to its original position.
    
    \item \textbf{Constraint Inference}: For each sub-task, we use LLM to systematically infer spatial and temporal constraints through a hierarchical constraint analysis process. This analysis begins with identifying the functional relationships between objects' key points and axes. For grasping sub-tasks, we derive constraints between the end-effector's pose and the object's annotated contact points and approach axis, ensuring stable and effective grasps. For manipulation sub-tasks, we establish geometric constraints between the tool's functional points and the target object. These constraints encompass both positional alignments and directional requirements.

    \item \textbf{Robot Behavior Generation}: Based on the derived spatial constraints, the LLM proceeds to generate corresponding behavioral code for each sub-task by calling relevant APIs (See prompts and examples in Appendix~\ref{appendix:prompts}). During execution, the system performs precise calculations of end-effector poses based on these spatial constraints.
The process begins by identifying functional points on the object within the world coordinate system, which serves as the fundamental reference frame for all subsequent pose calculations. Building upon this foundation, our system implements a dual approach to determine optimal target poses.
The first approach leverages pre-labeled contact points on the object to generate grasp poses. This method takes into account both the object's geometric properties and the robot's kinematic limitations. For more complex manipulation tasks, the second approach comes into play, computing target poses by aligning the object's functional point with a designated target point while adhering to specific directional constraints.
To illustrate this, consider a hammering task: the system would align the hammer's head with the nail while calculating the proper orientation for an effective strike. The core of behavior generation for each sub-task is an optimization problem that seeks optimal joint trajectories $\theta(t)$. Using a screw motion planner, the system minimizes a cost function $J(\theta(t))$ while satisfying all task-specific constraints. This optimization is formulated as:
\vspace{-3pt}
\begin{equation*}
\small
\begin{array}{ll}
\displaystyle \min_{\theta(t)} & J(\theta(t)) \\[2mm]
\text{s.t.} & 
\begin{cases}
\mathbf{T}_{\text{ee}} = f_{\text{FK}}(\theta(t)) \quad \text{(Kinematic constraint)} \\
\mathbf{P}_{\text{ee}} = \mathbf{P}_{o} - d \cdot \vec{a}_{o} \quad \text{(Position alignment)} \\
\vec{n}_{\text{ee}} = \vec{a}_{o} \quad \text{(Orientation alignment)} \\
\theta(t) \in \mathcal{C}, \forall t \in [t_0, t_f] \quad \text{(Collision avoidance)}
\end{cases}
\end{array}
\end{equation*}
where, \(J(\theta(t))\) represents a cost function that may incorporate factors such as energy efficiency, execution time, and motion smoothness. The constraints ensure that the robot's end-effector pose \(\mathbf{T}_{\text{ee}}\) matches the desired pose calculated through the forward kinematics function \(f_{\text{FK}}(\theta(t))\), aligning with the object's contact point \(\mathbf{P}_{o}\) and approach axis \(\vec{a}_{o}\) (position and orientation alignment). Finally, the trajectory \(\theta(t)\) must remain within the collision-free configuration space \(\mathcal{C}\) throughout the time interval \([t_0, t_f]\), ensuring collision avoidance. This comprehensive optimization framework enables the generation of robot behaviors that are efficient, satisfy spatial constraints, and guarantee safe, collision-free execution of complex tasks like hammering.

    
    
    \item \textbf{Success Evaluation}: We implement criteria to assess successful task completion. For the hammering task, this might include verifying that the nail has been driven to the correct depth.
    
    \item \textbf{Iterative Refinement}: The system gathers error data from multiple sources: runtime error messages, failed trajectory planning steps, and deviations between the final object states and their target configurations. To regenerate improved code, the system takes a comprehensive set of inputs including the collected error information, original task description, object annotations, and the previous version of code. The newly generated code is then tested, and if issues persist, the cycle continues until the desired performance is achieved. 
\end{enumerate}

\begin{figure*}[t]
    \centering
    \begin{minipage}[t]{0.6\textwidth}
        \centering
        \includegraphics[width=\textwidth]{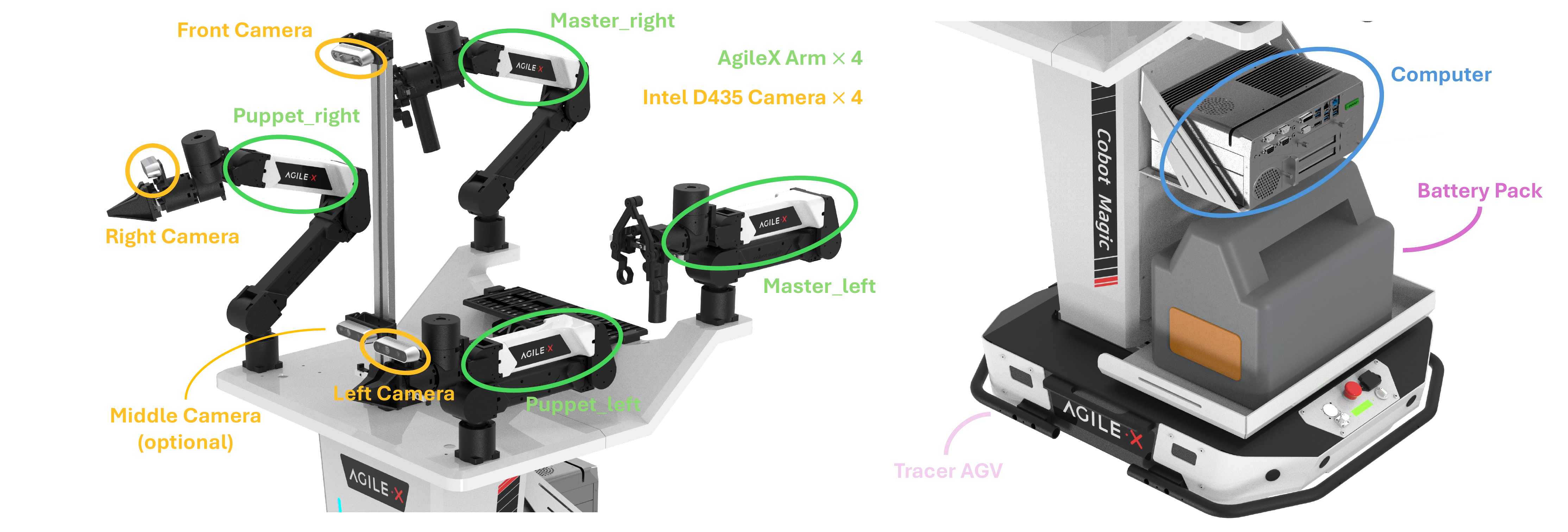}
        \captionof{figure}{\textbf{Illustration of our robot platform, with the capabilities for teleoperation and data acquisition.}}
        \label{fig:hardware1}
    \end{minipage}%
    \hfill
    \begin{minipage}[t]{0.35\textwidth}
        \centering
    \includegraphics[width=0.9\textwidth]{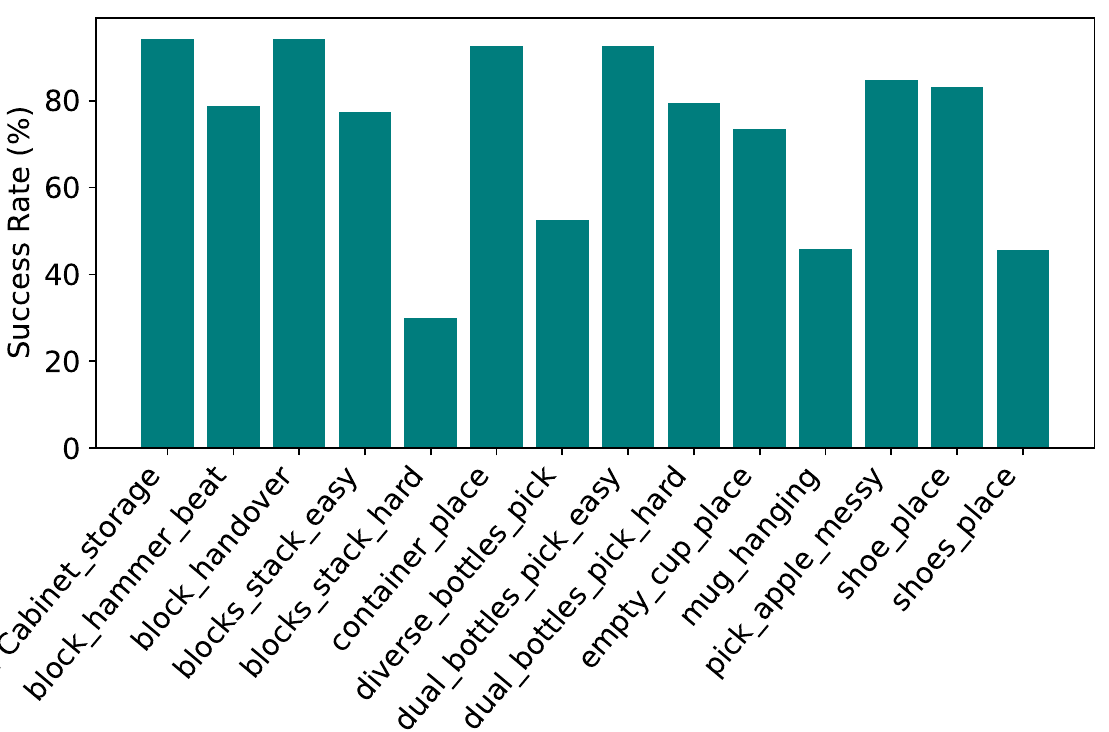}
        \captionof{figure}{\textbf{Success rate of the generated code for RoboTwin benchmark.}}
        \label{fig:code gen}
    \end{minipage}
    \vspace{-13pt}
\end{figure*}


\section{Benchmark}
\footnotetext[5]{Platform Introduction: \href{https://global.agilex.ai/products/cobot-magic}{https://global.agilex.ai/products/cobot-magic}}

Based on the methods introduced in Sec.~\ref{section:method}, we design a comprehensive benchmark called RoboTwin\cite{mu2024robotwin} to assess dual-arm robots, which includes 15 tasks in total. The underlying physics engine is ManiSkill3\cite{taomaniskill3}. We employ the open-source Cobot Magic\footnotemark[5] platform as depicted in Fig.~\ref{fig:hardware1}, which is equipped with four robot arms and four Intel RealSense D-435 RGBD cameras and is built on the Tracer chassis. These cameras are strategically positioned: one on the high part of the stand for an expansive field of view, two on the wrists of the robot's arms, and one on the low part of the stand which is optional for use. The front, left, and right cameras capture data simultaneously at a frequency of 30Hz. We utilize ManiSkill~\cite{taomaniskill3}, an open-source simulation platform with GPU-accelerated data collection built on SAPIEN~\cite{xiang2020sapien}. The details of each task in RoboTwin can be found in Appendix~\ref{appendix:task description}.

In RoboTwin benchmark, the agent needs to choose the appropriate collaboration method to successfully complete the task according to the distance of the target object from the left arm and the right arm. It involves the handover of the two arms, such as the handover task and putting the cup on the coaster, and the avoidance of interference between the two arms, such as the shoe placement task, which requires the two arms to coordinate with each other to place a pair of shoes in the limited space of the shoe box. The initial position and posture of the target objects in all our tasks are random. Before the scene is loaded, the mechanical dynamics accessibility of the randomly initialized scene will be checked to ensure that it is feasible. The task also includes objects of different shapes and appearances. The dual bottle pick task includes different models such as Coke bottles, Sprite bottles, and mineral water bottles, all of which are generated from 2D real pictures. The size of the objects in the environment is also randomized within a certain threshold. 
%
%
%
For each task, we provide well-designed script files that generate expert data across diverse scenarios, including various object placements and environmental conditions. We also report the success rate of generated code using our proposed method in Fig.~\ref{fig:code gen}, as described in Sec.~\ref{sec:code generation}.

For each task in our benchmark, we have pre-collected 100 sets of simulation data and 20 sets of real-world data. The hardware setup for the real-world experiments strictly matches that of the simulation environment. 
In both the simulation and real-world datasets, each captured frame consists of three images from the cameras, each providing an RGB and depth image. We also provide the point cloud data transformed from depth image, and colored point cloud data transformed from RGB and depth image for different types of algorithm evaluation.
Additionally, the data includes the poses of the robotic arms' joints and end-effectors for both master and slave configurations, encompassing both left and right arms. 

\begin{figure*}[t] 
    \centering    \includegraphics[width=0.85\textwidth]{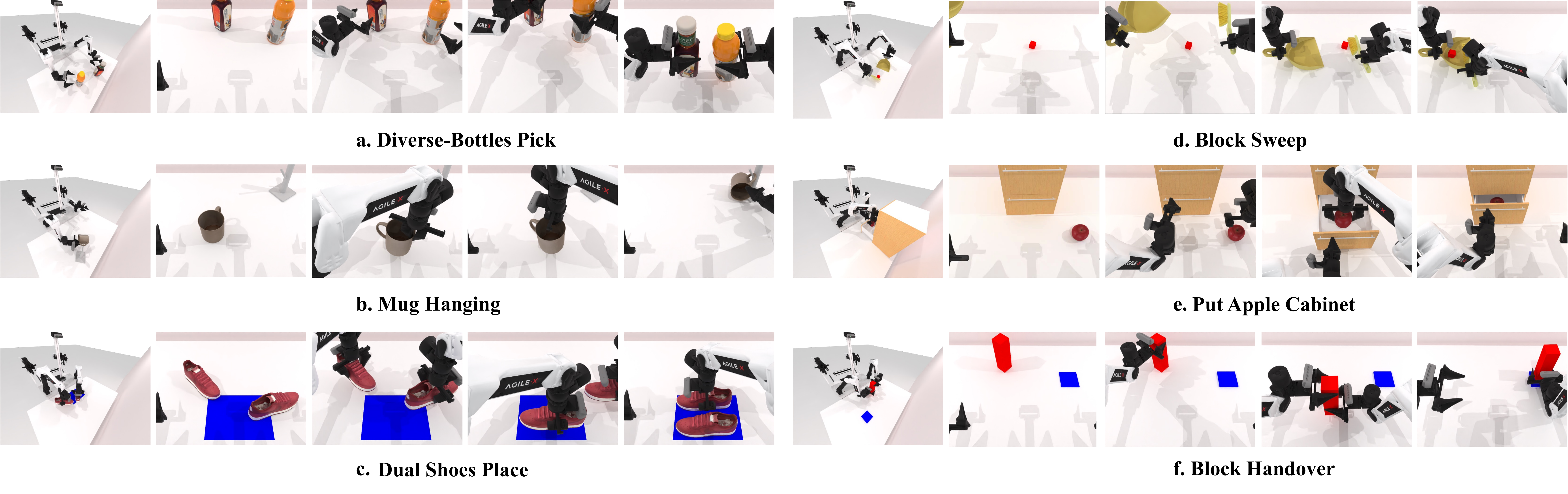}
    \vspace{-11pt}
    \caption{\textbf{Examples of task execution in the RoboTwin benchmark.}}
    \label{fig:benchmark_demo} 
    \vspace{-6pt}
\end{figure*}

\section{Experiment on RoboTwin Benchmark}

\begin{figure}[t] 
    \centering
\vspace{-7pt}
    \includegraphics[width=0.95\linewidth]{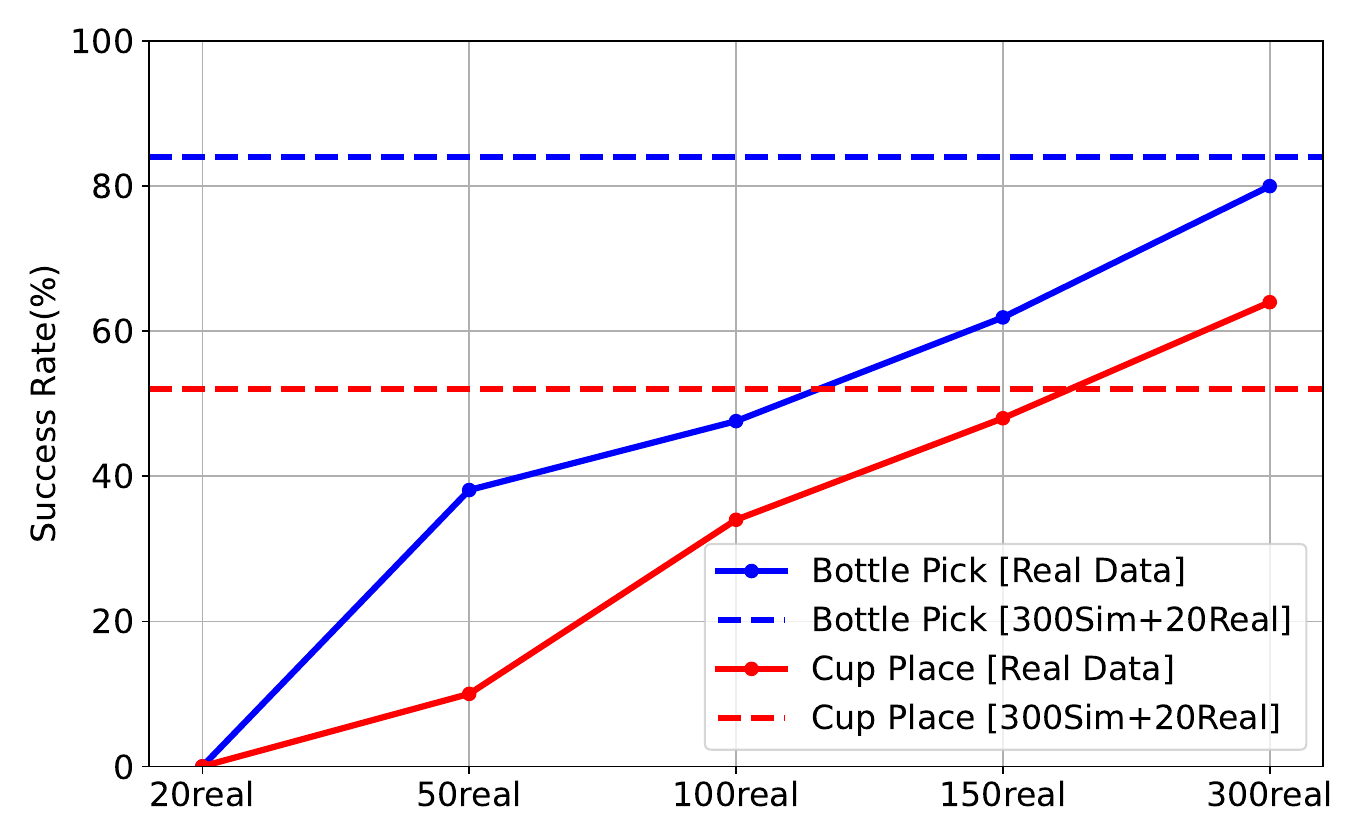}
    \vspace{-10pt}
    \caption{\textbf{Comparison on scaling up real and simulation data.}}
     \vspace{-18pt}
    \label{fig:scale_real}
\end{figure}

\subsection{Baselines and Experimental Setup}
Diffusion Policy is a generative model for robotic imitation learning that models the distribution of potential actions to create diverse and complex action sequences. The approach has evolved into two main variants based on input dimensionality:
The 2D Diffusion Policy~\cite{chi2023diffusion} processes two-dimensional visual information like images and video frames to predict actions for robotic manipulation tasks. While effective for many applications, this approach may have limitations in tasks requiring depth perception and spatial reasoning.
The 3D Diffusion Policy (DP3)\cite{ze20243d} addresses these limitations by incorporating three-dimensional visual representations through point clouds. By using efficient point encoders to create compact 3D representations, DP3 enhances spatial awareness and demonstrates improved performance in tasks requiring complex spatial understanding.
We evaluated both 3D (DP3, w \& w/o color) and 2D (DP) input imitation learning methods across 14 benchmark tasks, as shown in Fig.~\ref{fig:benchmark_demo}, tailoring our assessment approach to each model's characteristics using 20, 50, 100 expert demonstrations. The success rate is determined by satisfying the target pose constraints after execution completion and achieving collision-free trajectory execution throughout the task.

\subsection{Experimental Results}

\begin{table*}[t]
  \centering
  \small
\setlength{\tabcolsep}{10pt}
\resizebox{0.9\linewidth}{!}{
    \begin{tabular}{lccc|lccc}
    \toprule
       Number of Demonstrations   & \textbf{20} & \textbf{50} & \textbf{100} & & \textbf{20} & \textbf{50} & \textbf{100} \\
    \midrule
    \textbf{\textit{Block Hammer Beat}} & & & & \textbf{\textit{Block Handover}} & & & \\
    DP3 (XYZ) & 55.7 $\pm$ 8.5 & \bestcell{64.7 $\pm$ 10.1} & 55.7 $\pm$ 0.6 & DP3 (XYZ) & \bestcell{89.0 $\pm$ 2.6} & 84.3 $\pm$ 9.1 & 77.3 $\pm$ 11.6 \\
    DP3 (XYZ+RGB) & 47.7 $\pm$ 4.0 & 79.3 $\pm$ 3.8 & \bestcell{82.0 $\pm$ 6.6} & DP3 (XYZ+RGB) & 86.0 $\pm$ 1.0 & \bestcell{94.0 $\pm$ 0.0} & 85.3 $\pm$ 14.5 \\
    DP & 0.0 $\pm$ 0.0 & 0.0 $\pm$ 0.0 & 0.0 $\pm$ 0.0 & DP & 0.0 $\pm$ 0.0 & 12.0 $\pm$ 5.0 & \bestcell{76.0 $\pm$ 16.1} \\
    \midrule
    \textbf{\textit{Bottle Adjust}} & & & & \textbf{\textit{Container Place}} & & & \\
    DP3 (XYZ) & 64.7 $\pm$ 10.8 & 71.7 $\pm$ 13.8 & \bestcell{73.3 $\pm$ 12.5} & DP3 (XYZ) & 52.7 $\pm$ 5.0 & 77.7 $\pm$ 2.5 & \bestcell{85.3 $\pm$ 3.2} \\
    DP3 (XYZ+RGB) & 25.0 $\pm$ 5.0 & 36.0 $\pm$ 8.5 & \bestcell{42.0 $\pm$ 7.0} & DP3 (XYZ+RGB) & 37.3 $\pm$ 2.1 & 51.3 $\pm$ 7.1 & \bestcell{62.3 $\pm$ 6.8} \\
    DP & 6.3 $\pm$ 5.9 & 33.7 $\pm$ 9.0 & \bestcell{35.7 $\pm$ 2.9} & DP & 1.7 $\pm$ 0.6 & 8.0 $\pm$ 1.7 & \bestcell{14.0 $\pm$ 6.9} \\
    \midrule
    \textbf{\textit{Empty Cup Place}} & & & & \textbf{\textit{Mug Hanging (Easy)}} & & & \\
    DP3 (XYZ) & 33.7 $\pm$ 4.2 & \bestcell{71.3 $\pm$ 4.0} & 61.7 $\pm$ 13.1 & DP3 (XYZ) & 7.3 $\pm$ 3.2 & 14.0 $\pm$ 3.6 & \bestcell{15.3 $\pm$ 4.0} \\
    DP3 (XYZ+RGB) & 23.7 $\pm$ 5.5 & 68.0 $\pm$ 7.5 & \bestcell{81.0 $\pm$ 2.6} & DP3 (XYZ+RGB) & \bestcell{4.3 $\pm$ 3.1} & 1.7 $\pm$ 1.5 & 3.0 $\pm$ 1.0 \\
    DP & 0.0 $\pm$ 0.0 & 25.0 $\pm$ 2.6 & \bestcell{87.7 $\pm$ 0.6} & DP &  0.0 $\pm$ 0.0 & 0.0 $\pm$ 0.0 & 0.0 $\pm$ 0.0 \\
    \midrule
    \textbf{\textit{Mug Hanging (Hard)}} & & & & \textbf{\textit{Pick Apple Messy}} & & & \\
    DP3 (XYZ) & 4.0 $\pm$ 1.7 & 10.7 $\pm$ 3.1 & \bestcell{15.3 $\pm$ 5.5} & DP3 (XYZ) & 4.0 $\pm$ 1.7 & \bestcell{12.7 $\pm$ 5.5} & 9.7 $\pm$ 2.1 \\
    DP3 (XYZ+RGB) & 0.0 $\pm$ 0.0 & 1.7 $\pm$ 1.2 & \bestcell{2.3 $\pm$ 2.5} & DP3 (XYZ+RGB) & 6.0 $\pm$ 2.6 & 31.0 $\pm$ 7.5 & \bestcell{54.0 $\pm$ 12.8} \\
    DP & 0.0 $\pm$ 0.0 & 0.0 $\pm$ 0.0 & 0.0 $\pm$ 0.0 & DP & 5.3 $\pm$ 2.5 & 16.7 $\pm$ 1.5 & \bestcell{29.3 $\pm$ 5.0} \\
    \midrule
    \textbf{\textit{Put Apple Cabinet}} & & & & \textbf{\textit{Dual Bottles Pick (Easy)}} & & & \\
    DP3 (XYZ) & 50.0 $\pm$ 38.2 & \bestcell{73.3 $\pm$ 9.2} & 66.3 $\pm$ 22.3 & DP3 (XYZ) & 40.3 $\pm$ 8.0 & \bestcell{74.7 $\pm$ 2.9} & 55.3 $\pm$ 11.5 \\
    DP3 (XYZ+RGB) & 53.7 $\pm$ 14.2 &  54.3 $\pm$ 17.4  &  \bestcell{78.3 $\pm$ 3.8} & DP3 (XYZ+RGB) & 36.7 $\pm$ 5.9 & 74.7 $\pm$ 5.5 & \bestcell{75.7 $\pm$ 17} \\
    DP & 0.0 $\pm$ 0.0 & 0.0 $\pm$ 0.0 & \bestcell{8.0 $\pm$ 12.2} & DP & 1.7 $\pm$ 0.6 & 38.3 $\pm$ 6.7 & \bestcell{85.7 $\pm$ 6.7} \\
\midrule
\textbf{\textit{Dual Bottles Pick (Hard)}} & & & & \textbf{\textit{Diverse Bottles Pick}} & & & \\
    DP3 (XYZ) & 31.7 $\pm$ 9.0 & 48.0 $\pm$ 7.9 & \bestcell{58.0 $\pm$ 3.0} & DP3 (XYZ) & 11.3 $\pm$ 2.1 & 32.3 $\pm$ 10.1 & \bestcell{37.0 $\pm$ 10.0} \\
    DP3 (XYZ+RGB) & 28.0 $\pm$ 4.4 & 47.3 $\pm$ 4.2 & \bestcell{55.7 $\pm$ 4.9} & DP3 (XYZ+RGB) & 2.0 $\pm$ 1.0 & 7.7 $\pm$ 4.0 & \bestcell{14.7 $\pm$ 4.7} \\
    DP & 8.0 $\pm$ 2.0 & 39.3 $\pm$ 4.0 & \bestcell{59.3 $\pm$ 5.5} & DP &  0.7 $\pm$ 0.6 & 0.3 $\pm$ 0.6 & \bestcell{12.0 $\pm$ 5.3} \\
\midrule
\textbf{\textit{Shoe Place}} & & & & \textbf{\textit{Dual Shoes Place}} & & & \\
    DP3 (XYZ) & 38.0 $\pm$ 11.5 & \bestcell{59.3 $\pm$ 7.4} & 54.3 $\pm$ 0.6 & DP3 (XYZ) & 4.0 $\pm$ 1.0 & 7.7 $\pm$ 2.1 & \bestcell{12.0 $\pm$ 1.7} \\
    DP3 (XYZ+RGB) & 14.0 $\pm$ 2.6 & 44.3 $\pm$ 2.9 & \bestcell{54.0 $\pm$ 11.5} & DP3 (XYZ+RGB) & 1.7 $\pm$ 1.5 & 3.3 $\pm$ 0.6 & \bestcell{6.0 $\pm$ 1.0} \\
    DP & 3.0 $\pm$ 1.2 & 4.3 $\pm$ 3.2 & \bestcell{33.0 $\pm$ 15.8} & DP &  0.0 $\pm$ 0.0 & 1.7 $\pm$ 1.2 & \bestcell{3.0 $\pm$ 1.0} \\
    \bottomrule
    \end{tabular}}
\vspace{-6pt}
    \caption{\textbf{Benchmarking imitation learning algorithms for dual-arm manipulation under D435 camera setting.} We tested on 14 tasks with 20, 50, and 100 expert demonstrations on DP3 (XYZ), DP3 (XYZ+RGB), and DP with 3 seeds and reported the success rate.
}
  \label{tab:main-results}%
  \vspace{-10pt}
\end{table*}

As shown in Table \ref{tab:main-results}, the experimental results reveal distinct performance patterns across different imitation learning methods. DP3 demonstrates superior few-shot learning capabilities, achieving remarkable performance with merely 20 demonstrations. However, its performance exhibits limited scalability, with minimal improvements or even decreases as training data expands to 100 samples. 
Conversely, the DP algorithm shows poor initial performance with limited data, likely due to insufficient geometric priors, but demonstrates significant scalability as training samples increase. With 100 demonstrations, DP outperforms DP3 in several tasks, significantly improving from 1. 7\% to 85.7\% in the Dual Bottles Pick (Easy) task. This indicates superior learning capabilities with larger datasets. 
%
The integration of RGB data with point cloud representations yields inconsistent benefits, highlighting a fundamental limitation in current bimanual manipulation approaches. While DP3(XYZ+RGB) shows dramatic improvements in cluttered environments such as Pick Apple Messy, it simultaneously exhibits performance degradation in some other tasks like Container Place.  This indicates that better fusion representations of RGB semantic information and point cloud 3D information need to be developed (see more results in Appendix Table \ref{tab:main-results-L515}).

Experimental results show significant performance variation based on coordination complexity. Simple operations like Dual Bottles Pick achieved high success rates (85.7\% with DP at 100 demonstrations), while tasks requiring complex bimanual coordination, such as Dual Shoes Place, performed poorly (below 15\% success across all methods). Notably, tasks demanding complex dual-arm coordination significantly underperformed compared to those where robot arms could operate more independently, with arm selection based primarily on proximity to target objects. This highlights the current limitations in dual-arm coordination within imitation learning algorithms.

\subsection{Real World Experiment}

\begin{figure}[t] 
    \centering    \includegraphics[width=0.5\textwidth]{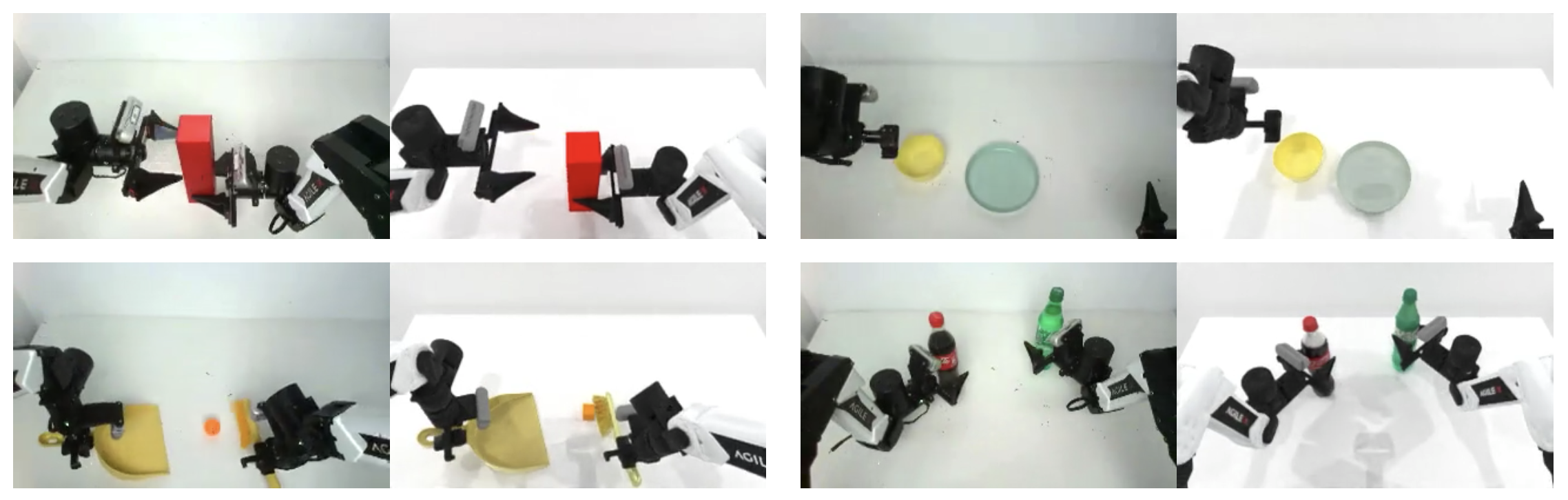}
    \vspace{-20pt}
    \caption{\textbf{Visualization of Real Scene and Simulation Scene.} More details can be found in Appendix Fig.~\ref{fig: Real world visualization}.}
    \label{fig:real-sim-demo} 
    \vspace{-8pt}
\end{figure}

To validate the effectiveness of RoboTwin-generated training data in real-world policy deployment, we conducted comprehensive experiments on both single-arm and dual-arm manipulation tasks, as shown in Fig. ~\ref{fig:real-sim-demo}.
We conducted a comparative experiment between policies trained solely on 20 real-world datasets and those pre-trained on 300 simulation datasets before fine-tuning on 20 real-world datasets (see more details and results in Appendix \ref{appendix:detail and results}).

The selection of 300 simulation datasets as our hyper-parameter was based on empirical evidence shown in Fig.~\ref{fig:scale_real}. Through progressive scaling of real-world data, we found that combining 300 simulation datasets with 20 real-world datasets yielded comparable performance than using 300 real-world datasets alone for both single-arm bottle pick and dual-arm cup placement tasks.

\begin{table}[t]
\small
\centering
\setlength{\tabcolsep}{13pt}
\resizebox{0.9\linewidth}{!}{
\begin{tabular}{c|c|c}
\toprule
 & \multicolumn{2}{c}{\textbf{Success Rates}} \\
 \midrule
\textbf{Task} & \textbf{20 real} & \textbf{300Sim+20Real} \\
\midrule
Bottle Pick (Easy) & 0/50 & \textbf{42/50} \\ 
Bottle Pick (Hard) & 0/50 & \textbf{16/50} \\ 
Container Place & 0/50 & \textbf{49/50} \\
Cup Place & 1/50 & \textbf{39/50} \\ 
Hammer Beat & 2/50 & \textbf{37/50} \\
\midrule
Average& 1.2\% & \textbf{72\%} \\ 
\bottomrule
\end{tabular}}
\vspace{-8pt}
\caption{\textbf{Real world evaluation with a single arm.}}
\label{tab:2d diffusion_policy single}
\vspace{-16pt}
\end{table}

To investigate the performance disparity between baseline algorithms in single-arm versus dual-arm tasks, we conducted sim-to-real transfer experiments for both task categories. Each task underwent 50 test trials with randomized initial configurations, including varying object positions and orientations, as well as robot arm placements within predetermined boundaries. As shown in Table~\ref{tab:2d diffusion_policy single} and Table~\ref{tab:2d diffusion_policy dual}, experimental results revealed that policies trained on the combined dataset achieved markedly superior performance in real-world testing scenarios. Specifically, the integration of simulation data yielded a 72\% improvement in success rates for single-arm tasks compared to policies trained exclusively on real-world data. For the more complex dual-arm tasks, we observed a significant improvement of over 40\% in success rates.
Our findings validate the effectiveness of our benchmark and data generation approach in bridging the sim-to-real gap, suggesting a promising direction for developing more robust and generalizable policies for dual-arm robotic manipulation tasks.


We observed significant disparities between single-arm and dual-arm scenarios. In the bottle rearrangement task, dual-arm operations presented substantially greater challenges, primarily due to the diverse initial states of target bottles (upright or lying down). While the incorporation of simulation data enabled the policy to achieve non-zero success rates, the overall performance remained suboptimal. This underscores the pressing need for developing more effective imitation learning algorithms specifically tailored to dual-arm coordination tasks.


\begin{table}[t]
\centering
\small
\setlength{\tabcolsep}{10pt}
\resizebox{0.9\linewidth}{!}{
\begin{tabular}{c|c|c}
\toprule
 & \multicolumn{2}{c}{\textbf{Success Rates}} \\
 \midrule
\textbf{Task} & \textbf{20 real} & \textbf{300Sim+20Real} \\
\midrule
Dual bottle Pick (Easy) & 0/50 & \textbf{31/50} \\ 
Dual bottle Pick (Hard) & 0/50 & \textbf{11/50} \\
Container Place &25/50 & \textbf{44/50} \\
Cup Place & 0/50 & \textbf{26/50} \\ 
Sweep Ball & 25/50 & \textbf{43/50} \\
\midrule
Average& 20\% & \textbf{62\%} \\ 
\bottomrule
\end{tabular}}
\vspace{-5pt}
\caption{\textbf{Real world evaluation with dual arms.}}
\label{tab:2d diffusion_policy dual}
\vspace{-15pt}
\end{table}


\section{Conclusion}
This work introduces RoboTwin, a comprehensive benchmark integrating real-world and synthetic data for dual-arm robotic manipulation.
Building upon the  COBOT Magic Robot platform and leveraging 3D generative models for generative digital twins, our framework enables the efficient generation of diverse training data from single RGB images.
Furthermore, our spatial-aware code generation framework automatically produces expert demonstrations by combining object annotations with LLMs to decompose complex tasks and generate precise movements. Experiments show that policies trained with RoboTwin-simulated data achieve higher success rates with less real data compared to those trained solely on real-world data. These results confirm our approach effectively bridges the sim-to-real gap while identifying limitations in dual-arm coordination tasks. Future work will focus on developing advanced algorithms for dual-arm coordination and expanding the framework to handle more complex manipulation tasks.

\clearpage
\section*{Acknowledgements}
We extend our profound gratitude to D-robotics for their invaluable support in supplying the necessary cloud computing resources that facilitated the execution of this research. 
Furthermore, we extend sincere appreciation to Deeoms for their contribution in providing essential model support, which was pivotal to the successful completion of this study. 
This paper is partially supported by the National Key R\&D Program of China No.2022ZD0161000 and the General Research Fund of Hong Kong No.17200622 and 17209324.
{
    \small
    \bibliographystyle{ieeenat_fullname}
    \bibliography{main}
}

\clearpage
\appendix
\setcounter{page}{1}
\maketitlesupplementary

\begin{table*}[t]
  \centering
  \small
\setlength{\tabcolsep}{10pt}
\resizebox{0.95\linewidth}{!}{
    \begin{tabular}{lccc|lccc}
    \toprule
       Number of Demonstrations   & \textbf{20} & \textbf{50} & \textbf{100} & & \textbf{20} & \textbf{50} & \textbf{100} \\
    \midrule
    \textbf{\textit{Block Hammer Beat}} & & & & \textbf{\textit{Block Handover}} & & & \\
    DP3 (XYZ) & 47.7 $\pm$ 7.4 & \bestcell{58.3 $\pm$ 6.5} & 49.7 $\pm$ 8.1 & DP3 (XYZ) & 82.7 $\pm$ 6.1 & \bestcell{85.0 $\pm$ 15.6} & 67.3 $\pm$ 7.0 \\
    DP3 (XYZ+RGB) & 44.7 $\pm$ 3.8 & \bestcell{79.0 $\pm$ 2.0} & 77.3 $\pm$ 7.5 & DP3 (XYZ+RGB) & 88.7 $\pm$ 5.0 & \bestcell{94.3 $\pm$ 7.2} & 86.0 $\pm$ 15.1\\
    DP & 0.0 $\pm$ 0.0 & 0.0 $\pm$ 0.0 & 0.0 $\pm$ 0.0 & DP & 0.0 $\pm$ 0.0 & 0.0 $\pm$ 0.0 & \bestcell{0.7 $\pm$ 1.2} \\
    \midrule
    \textbf{\textit{Bottle Adjust}} & & & & \textbf{\textit{Container Place}} & & & \\
    DP3 (XYZ) & 55.7 $\pm$ 1.5 & 70.7 $\pm$ 2.5 & \bestcell{72.7 $\pm$ 10.1} & DP3 (XYZ) & 52.7 $\pm$ 4.5 & 74.0 $\pm$ 5.6 & \bestcell{89.0 $\pm$ 7.5} \\
    DP3 (XYZ+RGB) & 28.3 $\pm$ 12.9 & 27.7 $\pm$ 16.5 & \bestcell{35.7 $\pm$ 12.5} & DP3 (XYZ+RGB) &38.0 $\pm$ 7.9 & 58.3 $\pm$ 5.9 & \bestcell{73.3 $\pm$ 6.5} \\
    DP & 13.0 $\pm$ 11.8 & 24.7 $\pm$ 13.8 & \bestcell{31.0 $\pm$ 6.6} & DP & 5.3 $\pm$ 4.2 & 16.3 $\pm$ 2.5 & \bestcell{35.0 $\pm$ 4.4} \\
    \midrule
    \textbf{\textit{Empty Cup Place}} & & & & \textbf{\textit{Mug Hanging (Easy)}} & & & \\
    DP3 (XYZ) & 33.0 $\pm$ 6.2 & 70.3 $\pm$ 7.2 & \bestcell{71.3 $\pm$ 20.4} & DP3 (XYZ) & 7.3 $\pm$ 2.9 & 14.0 $\pm$ 3.6 & \bestcell{14.7 $\pm$ 3.5} \\
    DP3 (XYZ+RGB) & 26.3 $\pm$ 10.4 & 71.3 $\pm$ 4.0 & \bestcell{78.7 $\pm$ 7.4} & DP3 (XYZ+RGB) & 1.0 $\pm$ 1.0 & \bestcell{2.0 $\pm$ 2.0} & \bestcell{2.0 $\pm$ 3.5} \\
    DP & 0.3 $\pm$ 0.6 & 14.7 $\pm$ 6.0 & \bestcell{58.0 $\pm$ 11.8} & DP & 0.0 $\pm$ 0.0 & 0.0 $\pm$ 0.0 & 0.0 $\pm$ 0.0 \\
    \midrule
    \textbf{\textit{Mug Hanging (Hard)}} & & & & \textbf{\textit{Pick Apple Messy}} & & & \\
    DP3 (XYZ) & \bestcell{12.7 $\pm$ 0.6} & 11.0 $\pm$ 6.1 & \bestcell{12.7 $\pm$ 2.3} & DP3 (XYZ) & 5.7 $\pm$ 4.5 & 10.7 $\pm$ 4.0 & \bestcell{11.7 $\pm$ 5.5} \\
    DP3 (XYZ+RGB) & 0.0 $\pm$ 0.0 & \bestcell{2.0 $\pm$ 2.0} & 0.3 $\pm$ 0.6 & DP3 (XYZ+RGB) & 6.7 $\pm$ 2.3 & 28.7 $\pm$ 9.5 & \bestcell{68.7 $\pm$ 6.8} \\
    DP & 0.0 $\pm$ 0.0 & 0.3 $\pm$ 0.6 & 0.0 $\pm$ 0.0 & DP & 3.3 $\pm$ 1.5 & 6.0 $\pm$ 5.0 & \bestcell{7.0 $\pm$ 4.6} \\
    \midrule
    \textbf{\textit{Put Apple Cabinet}} & & & & \textbf{\textit{Dual Bottles Pick (Easy)}} & & & \\
    DP3 (XYZ) & 60.7 $\pm$ 23.0 & \bestcell{89.3 $\pm$ 10.8} & 74.7 $\pm$ 42.2 & DP3 (XYZ) & 37.0 $\pm$ 4.6 & \bestcell{60.3 $\pm$ 7.1} & 32.0 $\pm$ 4.6 \\
    DP3 (XYZ+RGB) & 5.7 $\pm$ 4.0 & 96.0 $\pm$ 3.5 & \bestcell{97.0 $\pm$ 2.6} & DP3 (XYZ+RGB) & 29.7 $\pm$ 3.5 & 67.3 $\pm$ 9.3 & \bestcell{69.0 $\pm$ 23.5} \\
    DP & 1.3 $\pm$ 1.2 & 8.3 $\pm$ 2.5 & \bestcell{34.0 $\pm$ 21.2} & DP & 1.3 $\pm$ 1.5 & 26.7 $\pm$ 3.1 & \bestcell{79.0 $\pm$ 3.5} \\
    \midrule
    \textbf{\textit{Dual Bottles Pick (Hard)}} & & & & \textbf{\textit{Diverse Bottles Pick}} & & & \\
    DP3 (XYZ) & 33.0 $\pm$ 2.6 & 48.0 $\pm$ 5.2 & \bestcell{57.3 $\pm$ 4.0} & DP3 (XYZ) & 13.3 $\pm$ 5.5 & \bestcell{34.7 $\pm$ 6.7} & 33.7 $\pm$ 5.9 \\
    DP3 (XYZ+RGB) & 23.0 $\pm$ 2.0 & 46.3 $\pm$ 7.8 & \bestcell{56.7 $\pm$ 3.5} & DP3 (XYZ+RGB) & 0.7 $\pm$ 0.6 & 5.3 $\pm$ 2.1 & \bestcell{9.7 $\pm$ 2.9} \\
    DP & 2.0 $\pm$ 1.7 & 32.3 $\pm$ 5.9 & \bestcell{51.7 $\pm$ 5.1} & DP & 0.0 $\pm$ 0.0 & 0.3 $\pm$ 0.6 & \bestcell{6.0 $\pm$ 1.0} \\
    \midrule
    \textbf{\textit{Shoe Place}} & & & & \textbf{\textit{Dual Shoes Place}} & & & \\
    DP3 (XYZ) & 37.0 $\pm$ 10.5 & \bestcell{65.7 $\pm$ 11.5} & 54.0 $\pm$ 10.4 & DP3 (XYZ) & 5.7 $\pm$ 0.6 & 10.0 $\pm$ 2.6 & \bestcell{12.0 $\pm$ 2.0} \\
    DP3 (XYZ+RGB) & 19.7 $\pm$ 6.4 & 44.7 $\pm$ 4.0 & \bestcell{54.3 $\pm$ 2.5} & DP3 (XYZ+RGB) & 1.7 $\pm$ 2.9 & 3.7 $\pm$ 0.6 & \bestcell{7.7 $\pm$ 2.1} \\
    DP & 0.0 $\pm$ 0.0 & 6.3 $\pm$ 2.5 & \bestcell{27.0 $\pm$ 16.1} & DP &  0.0 $\pm$ 0.0 & 3.0 $\pm$ 1.7 & \bestcell{5.3 $\pm$ 2.9} \\
    \bottomrule
    \end{tabular}}
    \caption{\textbf{Benchmarking imitation learning algorithms for dual-arm manipulation under L515 camera setting.} We tested on 14 tasks with 20, 50, and 100 expert demonstrations on DP3 (XYZ), DP3 (XYZ+RGB), and DP, and reported the success rate and standard deviation.}
  \label{tab:main-results-L515}%

\end{table*}



\begin{figure*}[htbp]
    \centering 
    \includegraphics[width=0.95\textwidth]{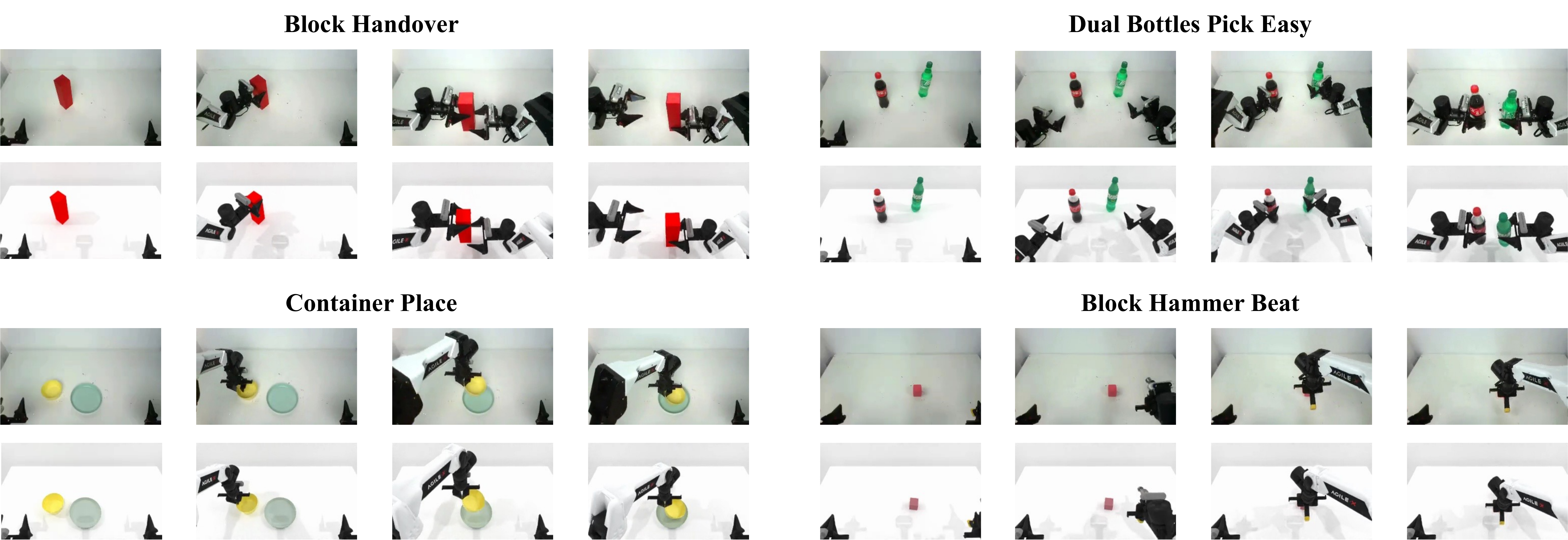}
    \caption{\textbf{Visualization of real-world and RoboTwin-generated data.} For each task, real-world collected data is shown in the top row, with RoboTwin-generated data displayed in the bottom row.}
    \label{fig: Real world visualization} 
\end{figure*}

\section{Task Description for RoboTwin}
\label{appendix:task description}
We provide detailed descriptions of all tasks involved in the benchmarks and real-world experiments, as shown in Table \ref{tab:benchmark_description}, totaling 15 tasks. The initial positions of target objects in all tasks are randomized. Some tasks must be completed using both arms, such as \textit{Shoes Place}. Other tasks have both dual-arm and single-arm versions, like \textit{Container Place} and \textit{Empty Cup Place}. For these dual-arm versions, the appropriate arm is selected based on the object's initial position. Tasks like \textit{Block Handover} and \textit{Mug Hanging} involve handoffs between the left and right arms. More challenging tasks, such as \textit{Shoes Place}, require high coordination between both arms.
\begin{table*}[ht]
\centering

\scriptsize

\begin{tabular}{>{\itshape\centering}m{3.5cm} p{8cm}} 
\toprule
\textbf{Task}   & \textbf{Description}        \\ \midrule 
Block Hammer Beat & There is a hammer and a block in the middle of the table. If the block is closer to the left robotic arm, it uses the left arm to pick up the hammer and strike the block; otherwise, it does the opposite. \\ \midrule
Block Handover & A long block is placed on the left side of the table. The left arm grasps the upper side of the block and then hands it over to the right arm, which places the block on the blue mat on the right side of the table. \\ \midrule
Bottle Adjust & A bottle is placed horizontally on the table. The bottle's design is random and does not repeat in the training and testing sets. When the bottle's head is facing left, pick up the bottle with the right robot arm so that the bottle's head is facing up; otherwise, do the opposite. \\ \midrule
Container Place & Random containers (cups, bowls, etc.) are placed randomly on the table. The robotic arm moves the containers into a fixed plate. \\ \midrule
Diverse Bottles Pick & A random bottle is placed on the left and right sides of the table. The bottles' designs are random and do not repeat in the training and testing sets. Both left and right arms are used to lift the two bottles to a designated location. \\ \midrule
Dual Bottles Pick (Easy) & A red bottle is placed randomly on the left side, and a green bottle is placed randomly on the right side of the table. Both bottles are standing upright. The left and right arms are used simultaneously to lift the two bottles to a designated location. \\ 
\midrule
Dual Bottles Pick (Hard) & A red bottle is placed randomly on the left side, and a green bottle is placed randomly on the right side of the table. The bottles' postures are random. Both left and right arms are used simultaneously to lift the two bottles to a designated location. \\ \midrule
Dual Shoes Place & One shoe is placed randomly on the left and right sides of the table. The shoes are the same pair with random designs that do not repeat in the training and testing sets. Both left and right arms are used to pick up the shoes and place them in the blue area, with the shoe heads facing the left side of the table. \\ \midrule
Empty Cup Place & An empty cup and a cup mat are placed randomly on the left or right side of the table. The robotic arm places the empty cup on the cup mat. \\ \midrule
Mug Hanging (Easy) & A mug is placed randomly on the left side of the table, and a mug rack is placed on the right side (fixed). The left arm moves the mug to a suitable position in the middle of the table, and then the right arm hangs the handle of the mug on the mug rack. \\ \midrule
Mug Hanging (Hard) & A mug is placed randomly on the left side of the table, and a mug rack is placed randomly on the right side. The left arm moves the mug to a suitable position in the middle of the table, and then the right arm hangs the handle of the mug on the mug rack. \\ \midrule
Pick Apple Messy & Apples and four random items are placed randomly on the table. The robotic arm picks up the apple and lifts it. \\  \midrule
Put Apple Cabinet & Initially, an apple is placed randomly. The robotic arm uses the left arm to open the cabinet and the right arm to pick up the apple and place them inside. \\ 
\midrule
Shoe Place & Shoes are placed randomly on the table, with random designs that do not repeat in the training and testing sets. The robotic arm moves the shoes to a blue area in the center of the table, with the shoe head facing the left side of the table. \\ 
\bottomrule
\end{tabular}
\caption{Task descriptions for RoboTwin platform.}
\label{tab:benchmark_description}
\end{table*}


\section{Implementation Details for Simulation Experiments}
\label{appendix:detail and results}
\subsection{Baseline Introduction and Setup}
Diffusion Policy~\cite{chi2023diffusion} is a novel approach in robot learning that models the robot's visuomotor policy as a conditional denoising diffusion process.  It learns the gradient of the action-distribution score function and iteratively optimizes with respect to this gradient field during inference via a series of stochastic Langevin dynamics steps. This methodology enables the robot to generate diverse and high-dimensional action distributions, effectively handling multi-modal behaviors and high-dimensional action spaces. The input to the Diffusion Policy is a sequence of visual observations, and the output is a sequence of actions predicted over a fixed duration, facilitating robust and temporally consistent action generation. 

Building upon the Diffusion Policy, the 3D Diffusion Policy (DP3)~\cite{ze20243d} integrates 3D visual representations into the diffusion framework, enhancing the robot's ability to generalize across various tasks and environments. DP3 employs a compact 3D visual representation extracted from sparse point clouds using an efficient point encoder. The input to DP3 is a 3D scene representation, and the output is a sequence of 3D end-effector poses, including both translations and rotations, predicted over a fixed duration. This approach allows the robot to perform complex manipulation tasks with high precision and generalization capabilities, even with limited demonstrations.  

 We outline all the key hyper-parameters for DP~\cite{chi2023diffusion} and DP3~\cite{ze20243d} in Table \ref{tab:DP and DP3 Setting}. These hyper-parameters were adopted directly from the original DP and DP3 papers to ensure consistent performance and enable fair comparison with the published results. 

For the camera settings, we utilize a 2D observation with an image resolution of (320, 240) and perform FPS downsampling on the point cloud obtained from the image to 1024 points for 3D observation.




\section{Sim2Real Experiment Setup}
Our real-world experiments aim to verify whether the generated simulation data can effectively aid in policy learning, enabling high performance in real-world testing despite exposure to only limited real-world data.
\subsection{Simulation vs. Real Scene Visualization}
We present the comparison images of the real and simulation for the same task in Fig.~\ref{fig: Real world visualization}. The RoboTwin-generated data demonstrates exceptional visual fidelity to real-world scenarios across all tasks. The simulated environment achieves near photo-realistic quality, accurately capturing lighting, shadows, and object textures. This high-fidelity simulation shows great promise for robot learning by effectively bridging the sim-to-real gap.

\subsection{Details of Sim2Real Fine-Tuning}

To better align real-world and simulation images, and considering that brighter environments facilitate better policy learning and feature extraction, we enhanced the typically darker real-world observations. We applied the following brightness adjustment code, where the alpha parameter can be fine-tuned based on specific lighting conditions:

\begin{lstlisting}[language=Python, frame=none, basicstyle=\small\ttfamily, commentstyle=\color{blue}\small\ttfamily,columns=fullflexible, breaklines=true, postbreak=\mbox{\textcolor{red}{$\hookrightarrow$}\space}, escapeinside={(*}{*)}]
cv2.convertScaleAbs(src, alpha=1.5, beta=0)
\end{lstlisting}



\textbf{Step 1:} We pretrain a Diffusion Policy network using 300 sets of RoboTwin-generated simulation data. This simulation data provides a rich foundation for learning basic manipulation skills. The pretraining phase follows the hyperparameter settings detailed in Tab.~\ref{tab:finetune}.

\textbf{Step 2:} Following the pretraining phase, we implement a highly efficient fine-tuning approach using only 20 sets of real-world robot data. This minimal data requirement significantly reduces the burden of real-world data collection while still enabling effective domain adaptation. The fine-tuning process builds upon the pretrained policy network from Step 1, adjusting the network parameters to bridge the sim-to-real gap. All fine-tuning hyperparameters are carefully selected and documented in Tab.~\ref{tab:finetune} to ensure optimal transfer learning performance.

This two-stage training strategy effectively combines the advantages of abundant simulation data with minimal real-world data requirements, demonstrating an efficient approach to robot skill acquisition and transfer.

\begin{table*}[ht]
\centering
\small
\begin{tabular}{ccc}
\toprule
Parameter & DP~\cite{chi2023diffusion} & DP3~\cite{ze20243d} \\ 
\midrule
horizon & 8 & 8 \\ 
n\_obs\_steps & 3 & 3 \\ 
n\_action\_steps & 6 & 6 \\ 
num\_inference\_steps & 100 & 10 \\ 
dataloader.batch\_size & 128 & 256 \\ 
dataloader.num\_workers & 0 & 8 \\ 
dataloader.shuffle & True & True \\ 
dataloader.pin\_memory & True & True \\ 
dataloader.persistent\_workers & False & False \\ 
optimizer.\_target\_ & torch.optim.AdamW & torch.optim.AdamW \\ 
optimizer.lr & 1.0e-4 & 1.0e-4 \\ 
optimizer.betas & [0.95, 0.999] & [0.95, 0.999] \\ 
optimizer.eps & 1.0e-8 & 1.0e-8 \\ 
optimizer.weight\_decay & 1.0e-6 & 1.0e-6 \\ 
training.lr\_scheduler & cosine & cosine \\ 
training.lr\_warmup\_steps & 500 & 500 \\ 
training.num\_epochs & 300 & 3000 \\ 
training.gradient\_accumulate\_every & 1 & 1 \\ 
training.use\_ema & True & True \\ 
\bottomrule
\end{tabular}
\caption{Hyper-parameter Settings for Training and Deployment of DP and DP3 Algorithms.}
\label{tab:DP and DP3 Setting}
\end{table*}
\begin{table*}[ht]
\centering
\small
\begin{tabular}{ccc}
\toprule
Parameter & Pre-training & Fine-tuning \\ 
\midrule
horizon & 8 & 8 \\ 
n\_obs\_steps & 3 & 3 \\ 
n\_action\_steps & 6 & 6 \\ 
num\_inference\_steps & 100 & 100 \\ 
dataloader.batch\_size & 128 & 128 \\ 
dataloader.num\_workers & 0 & 0 \\ 
dataloader.shuffle & True & True \\ 
dataloader.pin\_memory & True & True \\ 
dataloader.persistent\_workers & False & False \\ 
optimizer.\_target\_ & torch.optim.AdamW & torch.optim.AdamW \\ 
optimizer.lr & 1.0e-4 & 5e-5 \\ 
optimizer.betas & [0.95, 0.999] & [0.95, 0.999] \\ 
optimizer.eps & 1.0e-8 & 1.0e-8 \\ 
optimizer.weight\_decay & 1.0e-6 & 1.0e-6 \\ 
training.lr\_scheduler & cosine & cosine \\ 
training.lr\_warmup\_steps & 500 & 500 \\ 
training.num\_epochs & 300 & 300 \\ 
training.gradient\_accumulate\_every & 1 & 1 \\ 
training.use\_ema & True & True \\ 
training.rollout\_every & 50 & 50 \\ 
\bottomrule
\end{tabular}
\caption{Hyper-parameter Settings for Pretraining with RoboTwin-generated Data and Finetuning with Limited Real-world Data.}
\label{tab:finetune}
\end{table*}


\clearpage
\onecolumn
\section{Prompts}
\label{appendix:prompts}
In the process of generating expert demonstration data, we structure prompts for large language models with three components: 1) Task Information and General Prompt; 2) Introduction to Available APIs, detailing usable programming interfaces and libraries; 3) Function Examples that demonstrate implementation patterns. 
\subsection{Task Information and General Prompt}
\begin{promptcode}[Blocks Stack Hard Task Prompt Example]
    You need to generate relevant code for some robot tasks in a robot simulation environment based on the provided API.
    In this environment, distance 1 indicates 1 meter long. Pose is representated as 7 dimention, [x, y, z, qw, qx, qy, qz]. For a 7-dimensional Pose object, you can use Pose.p to get the [x, y, z] coordinates and Pose.q to get the [qw, qx, qy, qz] quaternion orientation.
    All functions which has parameter actor_data, and all of actor_data should be in the actor_data_dic.
    In the world coordinate system, the positive directions of the xyz coordinate axes are right, front, and upper respectively, so the direction vectors on the right, front, and upper sides are [1,0,0], [0,1,0], [0,0,1] respectively. In the same way, we can get the unit vectors of the left side, back side and down side.

    Task Discription:
    Use the gripper to pick up block1 and move block 1 to the target position. Then pick up block 2 and place it on the block 1, and finally pick up block3 and place it on the block2. If block1's x coordinate (dim 0) is greater than 0, use right arm to stack the block1, else use the left arm. And same for the block2 and block3.
    Note: 
    1. You need to call the get_avoid_collision_pose function to avoid collisions when the left and right arms move alternately. 
    2. For example, if the previous action uses the left arm and the next action uses the right arm, you need to move the left arm after release gripper to avoid collisions, vice versa. 
    3. The pre-dis of stacked blocks may be smaller.

    Available Constants:
    self.world_direction_dic: {
        'left':         [0.5,  0.5,  0.5,  0.5],
        'front_left':   [0.65334811, 0.27043713, 0.65334811, 0.27043713],
        'front' :       [0.707, 0,    0.707, 0],
        'front_right':  [0.65334811, -0.27043713,  0.65334811, -0.27043713],
        'right':        [0.5,    -0.5, 0.5,  0.5],
        'top_down':     [0,      0,   1,    0],
    }
    The world_direction_dic is a dict of different approach directions.
    The Actor Name List: ['block1', 'block2', 'block3', 'block1_target_pose']
    The Actor Data List: ['block1_data', 'block2_data', 'block3_data', 'block1_target_pose']

    The Actor Points Discription: {
        'block1':{
            'contact_points':[]
            'target_points': ["The top surface center of the block." ],
            'functional_points': ["Point0: The center point on the bottom of the block, and functional axis is vertical bottom side down"]
            'actor_orientation': []
        },
        'block2':{
            'contact_points':[]
            'target_points': ["The top surface center of the block." ],
            'functional_points': ["Point0: The center point on the bottom of the block, and functional axis is vertical bottom side down"]
            'actor_orientation': []
        },
        'block3':{
            'contact_points':[]
            'target_points': ["The top surface center of the block." ],
            'functional_points': ["Point0: The center point on the bottom of the block, and functional axis is vertical bottom side down"]
            'actor_orientation': []
        }
    }

    Current Code:
    ```python
    class gpt_{dual_bottles_pick_hard}({dual_bottles_pick_hard}):
        def play_once(self):
            pass
    ```
\end{promptcode}

\subsection{Introduction of Available APIs}
\begin{promptcode}
    Available API:
    "open_left_gripper": Open the left gripper to a specified position.,
    "close_left_gripper": Close the left gripper to a specified position.,
    "open_right_gripper": Open the right gripper to a specified position.,
    "close_right_gripper": Close the right gripper to a specified position.,
    "together_open_gripper": Open both left and right grippers to specified positions.,
    "together_close_gripper": Close both left and right grippers to specified positions.,

    "left_move_to_pose_with_screw": 
        def left_move_to_pose_with_screw(pose).
        Plan and execute a motion for the left arm using screw motion interpolation.
        No Return.
        Args:
        pose: list [x, y, z, qw, qx, qy, qz], the target pose of left end-effector,
    "right_move_to_pose_with_screw": 
        def right_move_to_pose_with_screw(pose).
        Plan and execute a motion for the right arm using screw motion interpolation.
        No Return.
        Args:
        pose: list [x, y, z, qw, qx, qy, qz], the target pose of right end-effector,
    "together_move_to_pose_with_screw": 
        def together_move_to_pose_with_screw(left_target_pose, right_target_pose).
        Plan and execute motions for both left and right arms using screw motion interpolation.
        No Return.
        Args:
        left_target_pose: list [x, y, z, qw, qx, qy, qz], the target pose of left end-effector
        right_target_pose: list [x, y, z, qw, qx, qy, qz], the target pose of right end-effector,

    "get_actor_functional_pose":
        def get_actor_functional_pose(actor, actor_data),
        Get the functional pose of the actor in the world coordinate system.
        Returns: pose: list [x, y, z, qw, qx, qy, qz].
        Args:
        actor: Object(self.actor), the object of actor in render.
        actor_data: dict(self.actor_data), the actor_data match with actor.,

    "get_grasp_pose_to_grasp_object":
        def get_grasp_pose_to_grasp_object(self, endpose_tag: str, actor, actor_data = DEFAULT_ACTOR_DATA, pre_dis = 0),
        This function is used to grasp actor from the labeled contact points of the actor, and return the most suitable pose of the end-effector.
        Returns: pose: list [x, y, z, qw, qx, qy, qz].
        Args:
            endpose_tag: str, the endpose tag of the actor, can be 'left' or 'right'.
            actor: Object(self.actor), the object of actor in render.
            actor_data: dict(self.actor_data), the actor_data match with actor.
            pre_dis: float, the distance between grasp pose and target actor pose.,

    "get_grasp_pose_from_goal_point_and_direction": 
        def get_grasp_pose_from_goal_point_and_direction(self, actor, actor_data,  endpose_tag: str, actor_functional_point_id, target_point, target_approach_direction, actor_target_orientation = None, pre_dis):
        This function is used to move the actor's point of action to the target point when the direction of the end-effector is given, return the pose of the end-effector.
        The actor refers to an object being grasped by robotic grippers. actor_target_orientation is the orientation of the actor after grasping.
        Returns: pose: list [x, y, z, qw, qx, qy, qz].
        Args: 
        actor: Object(self.actor), the object of actor in render.
        actor_data: dict(self.actor_data), the actor_data match with actor.
        endpose_tag: str, the endpose tag of the actor, can be 'left' or 'right'.
        actor_functional_point_id: int, the index of the functional point of the actor.
        target_point: list [x, y, z], the target point pose which the actor's target_pose expected to move to.
        target_approach_direction: list [qw, qx, qy, qz], the approach direction which the actor's expected approach direction at the target point. 
        The target approach direction can use self.world_direction_dic['left', 'front_left', 'front', 'fron_right', 'right', 'top_down'].
        actor_target_orientation: list [x, y, z], the orientation of the actor after grasping. The positive directions of the xyz axis are right, front, and up respectively. You can give a direction vector to specify the target direction of the object. like [0, 0, 1] means the actor' orientation is up and [0, 1, 0] means the actor's orientation is front.
        pre_dis: float, the distance on approach direction between actor's point of action and target point.,

    "get_avoid_collision_pose":
        def get_avoid_collision_pose(self, avoid_collision_arm_tag: str),
        This function can obtain the safe position of the specified robot arm to avoid collision when both arms need to move at the same time.
        Returns: pose: list [x, y, z, qw, qx, qy, qz].
        Args:
        avoid_collision_arm_tag: str, 'left' or 'right'.,

    "get_actor_goal_pose":
        def get_actor_goal_pose(self, actor, actor_data, id),
        This function is used to get the target pose point of an actor in world axis.
        Returns: pose: list [x, y, z].
        Args:
        actor: Object(self.actor), the object of actor in render.
        actor_data: dict(self.actor_data), the actor_data match with actor.
        id: int, the id of the actor, if the actor has multiple target points. And default is 0.,
\end{promptcode}

\vspace{10pt}
\subsection{Function Example}
\begin{promptcode}
    Function Example:
    You can retrieve the actor object by the actor's name:
    ```python
    actor = self.actor_name_dic['actor_name']
    ```
    You can retrieve the actor_data object by the actor_data's name:
    ```python
    actor_data = self.actor_data_dic['actor_data_name']
    ```
    
    Here are some APIs and examples of grasping objects:
    If you want to get the gripper pose to grasp the actor, you typically execute the following code:
    ```python
    grasp_pose = self.get_grasp_pose_to_grasp_object(endpose_tag = "left", self.actor, self.actor_data, pre_dis = 0.09)  # endpose_tag can be "left" or "right"
    ```
    
    If you want to pick up an actor, you can refer to the following sample code:
    ```python
    pre_grasp_pose = self.get_grasp_pose_to_grasp_object(endpose_tag = "left", self.actor, self.actor_data, pre_dis = 0.09)  # endpose_tag can be "left" or "right"
    target_grasp_pose = self.get_grasp_pose_to_grasp_object(endpose_tag = "left", self.actor, self.actor_data, pre_dis = 0)  # endpose_tag can be "left" or "right"
    self.left_move_to_pose_with_screw(pre_grasp_pose)      # left arm move to the pre grasp pose
    self.left_move_to_pose_with_screw(target_grasp_pose)   # left arm move to the grasp pose
    self.close_left_gripper()                              # close left gripper to grasp the actor
    self.left_move_to_pose_with_screw(pre_grasp_pose)      # lift the actor up
    ```
    The code for grasping with the right arm or both arms is similar to the above code.
    
    For the grasping of a certain actor, the movement of the end-effector typically executes the following codes:
    ```python
    actor_pose = self.get_actor_goal_pose(self.actor, self.actor_data)
    
    if actor_pose[0] > 0:           # if the actor in the right side, use right arm to grasp the actor
        # grasp actor with right arm
    else:                           # if the actor in the left side, use left arm to grasp the actor
        # grasp actor with left arm
    ```
    
    Here are some examples of gripper control:
    ```python
    self.close_left_gripper(pos = 0.02)     # Close half of the left gripper
    self.close_left_gripper(pos = -0.01)    # Tighten the left gripper.
    self.open_left_gripper(pos = 0.02)      # Open half of the left gripper
    self.close_right_gripper(pos = 0.02)    # Close half of the right gripper
    self.close_right_gripper(pos = -0.01)   # Tighten the right gripper.
    self.open_right_gripper(pos = 0.02)     # Open half of the right gripper
    self.together_close_gripper(left_pos = 0.02,right_pose = 0.02) # Together close half of grippers
    ```
    Note:
    For grabbing some objects, you may need to close the clamping jaws tightly to grab them. You can adjust this through the 'pos' parameter, like 'pos = -0.01'.
    By default 'pos' is 0, when close gripper.
    
    Here are some APIs and examples of moving objects:
    Note: The drop height of the actor depends on the distance of the actor that was lifted up the previous action.
    To move an object to the target point, the 'get_grasp_pose_from_goal_point_and_direction()' is often called first to obtain the target's gripper posture.
    
    If you want to move the point of actor which is grasped by the gripper action to the target point, you typically execute the following code:
    ```python
    pre_grasp_pose = self.get_grasp_pose_from_goal_point_and_direction(self.actor, self.actor_data, endpose_tag = "left", actor_functional_point_id = 0, target_pose, target_approach_direction, pre_dis = 0.09)
    target_grasp_pose = self.get_grasp_pose_from_goal_point_and_direction(self.actor, self.actor_data, endpose_tag = "left", actor_functional_point_id = 0, target_pose, target_approach_direction, pre_dis = 0)
    self.left_move_to_pose_with_screw(pre_grasp_pose)      # left arm move to the pre grasp pose
    self.left_move_to_pose_with_screw(target_grasp_pose)   # left arm move to the grasp pose
    self.open_left_gripper()  # open left gripper to place the target object
    # You also can move right arm
    ```
    Note:
    1. The target_approach_direction is the approach direction which the actor's expected approach direction at the target point.
    2. actor_functional_point_id is the index of the functional point of the actor, You can choose based on the given function points information.
    3. For the parameter target_approach_direction, you can use self.world_direction_dic['left', 'front_left', 'front', 'fron_right', 'right', 'top_down'].
    4. The target pose can be obtained by calling the 'get_actor_goal_pose()' function.
    
    If you also have requirements for the target orientation of the object, you can specify the actor_target_orientation parameter through the direction vector to determine the final orientation of the object:
    ```python
    # the actor target orientation is front, the direction vector is [0,1,0]
    # The positive directions of the direction vector xyz axis are right, front, and up respectively.
    pre_grasp_pose = self.get_grasp_pose_from_goal_point_and_direction(self.actor, self.actor_data, endpose_tag = "left", actor_functional_point_id = 0, target_pose, actor_target_orientation = [0,1,0], target_approach_direction, pre_dis = 0.09)
    target_grasp_pose = self.get_grasp_pose_from_goal_point_and_direction(self.actor, self.actor_data, endpose_tag = "left", actor_functional_point_id = 0, target_pose, actor_target_orientation = [0,1,0], target_approach_direction, pre_dis = 0)
    self.left_move_to_pose_with_screw(pre_grasp_pose)      # left arm move to the pre grasp pose
    self.left_move_to_pose_with_screw(target_grasp_pose)   # left arm move to the grasp pose
    self.open_left_gripper()  # open left gripper to place the target object
    ```
    
    If you need to align the functional axis of the grabbed object with the functional axis of the target object, you can use the following code:
    ```python
    target_actor_functional_pose = self.get_actor_functional_pose(self.actor, self.actor_data, actor_functional_point_id = 0)
    target_actor_point = target_actor_functional_pose[:3]
    target_approach_direction = target_actor_functional_pose[3:]
    pre_grasp_pose = self.get_grasp_pose_from_goal_point_and_direction(self.actor, self.actor_data, endpose_tag = "left", actor_functional_point_id = 0, target_point = target_actor_point, target_approach_direction = target_approach_direction, pre_dis = 0.09)
    target_grasp_pose = self.get_grasp_pose_from_goal_point_and_direction(self.actor, self.actor_data, endpose_tag = "left", actor_functional_point_id = 0, target_point = target_actor_point, target_approach_direction = target_approach_direction, pre_dis = 0)
    self.left_move_to_pose_with_screw(pre_grasp_pose)      # left arm move to the pre grasp pose
    self.left_move_to_pose_with_screw(target_grasp_pose)   # left arm move to the grasp pose
    self.open_left_gripper()  # open left gripper to place the target object
    ```
    Note: 
    1. The parameter actor in get_grasp_pose_from_goal_point_and_direction() should be grasp actor, not the target actor.
    2. self.world_direction_dic is a dict of different approach directions.
    3. This situation usually occurs when hanging objects or performing some delicate operations.
    4. actor_functional_point_id is the index of the functional point of the actor, You can choose based on the given function points information.
    
    Some tasks involve simultaneous operations of the left and right arms, which may require calling the collision avoidance function:
    1. There is no need to avoid collision at the end of the task.
    2. If both arms have moved at the same time before, and the next step needs to be to move the left arm first to place the target object, You can first obtain the pose of the right arm that can avoid subsequent collisions, and then move both arms at the same time:
    ```python
    # Get left and right arm target pose
    # Here, the direction in which the object contacts the target point is vertically top_down as an example.
    # The actor target orientation is left, the direction vector is [-1,0,0].
    left_pre_pose = self.get_grasp_pose_from_goal_point_and_direction(left_actor, left_actor_data, endpose_tag="left", actor_functional_point_id = 0, target_point=point1, target_approach_direction=self.world_direction_dic['top_down'], actor_target_orientation=[-1, 0, 0], pre_dis=0.05)
    left_target_pose = self.get_grasp_pose_from_goal_point_and_direction(left_actor, left_actor_data, endpose_tag="left", actor_functional_point_id = 0, target_point=point1, target_approach_direction=self.world_direction_dic['top_down'], actor_target_orientation=[-1, 0, 0], pre_dis=0)
    right_pre_pose = self.get_grasp_pose_from_goal_point_and_direction(right_actor, right_actor_data, endpose_tag="right", actor_functional_point_id = 0, target_point=point2, target_approach_direction=self.world_direction_dic['top_down'], actor_target_orientation=[-1, 0, 0], pre_dis=0.05)
    right_target_pose = self.get_grasp_pose_from_goal_point_and_direction(right_actor, right_actor_data, endpose_tag="right", actor_functional_point_id = 0, target_point=point2, target_approach_direction=self.world_direction_dic['top_down'], actor_target_orientation=[-1, 0, 0], pre_dis=0)
    # right arm avoid collision pose
    right_avoid_collision_pose = self.get_avoid_collision_pose(avoid_collision_arm_tag = 'right')
    # move left arm to the pre pose and right arm to the avoid collision pose
    self.together_move_to_pose_with_screw(left_pre_pose, right_avoid_collision_pose)
    # put down the actor on left gripper
    self.left_move_to_pose_with_screw(left_target_pose)
    self.open_left_gripper()  # open left gripper to place the target object
    # left arm avoid collision pose
    left_avoid_collision_pose = self.get_avoid_collision_pose(avoid_collision_arm_tag = 'left')
    # move right arm to the target pose and left arm to the avoid collision pose
    self.together_move_to_pose_with_screw(left_avoid_collision_pose, right_pre_pose)
    # put down the actor on right gripper
    self.right_move_to_pose_with_screw(right_target_pose)
    self.open_right_gripper()  # open right gripper to place the target object
    # avoid_collision_arm_tag: 'left' or 'right'
    # direction: 'left', 'right', 'front', 'back', 'up', 'down'
    ```
    Note: 
    1. If the move_arm_tag is 'left', the direction also not be 'right', and same for 'right'.
    2. Collision avoidance may only be necessary if both arms have been moved.
\end{promptcode}

\vspace{10pt}
\subsection{Sample Code for Blocks Stack Hard Task}
\begin{pythoncode}[Sample of Play_once Function on Blocks Stack Hard Task]
    class gpt_blocks_stack_hard(blocks_stack_hard):
        def play_once(self):
            # Retrieve actor objects
            block1 = self.actor_name_dic['block1']
            block2 = self.actor_name_dic['block2']
            block3 = self.actor_name_dic['block3']
            block1_target_pose = self.actor_name_dic['block1_target_pose']

            # Retrieve actor data objects
            block1_data = self.actor_data_dic['block1_data']
            block2_data = self.actor_data_dic['block2_data']
            block3_data = self.actor_data_dic['block3_data']
            block1_target_pose_data = self.actor_data_dic['block1_target_pose']

            # Define pre-dis for grasping and placing
            pre_dis = 0.08

            # Function to grasp and place a block
            def grasp_and_place(block, block_data, target_pose, target_pose_data, pre_dis):
                # Determine which arm to use based on the block's x coordinate
                block_pose = self.get_actor_goal_pose(block, block_data)
                if block_pose[0] > 0:
                    arm_tag = "right"
                    move_function = self.right_move_to_pose_with_screw
                    close_gripper_function = self.close_right_gripper
                    open_gripper_function = self.open_right_gripper
                else:
                    arm_tag = "left"
                    move_function = self.left_move_to_pose_with_screw
                    close_gripper_function = self.close_left_gripper
                    open_gripper_function = self.open_left_gripper

                # Get the grasp pose
                pre_grasp_pose = self.get_grasp_pose_to_grasp_object(endpose_tag=arm_tag, actor=block, actor_data=block_data, pre_dis=pre_dis)
                target_grasp_pose = self.get_grasp_pose_to_grasp_object(endpose_tag=arm_tag, actor=block, actor_data=block_data, pre_dis=0)

                # Move to the pre-grasp pose
                move_function(pre_grasp_pose)

                # Move to the grasp pose
                move_function(target_grasp_pose)

                # Close the gripper to grasp the block
                close_gripper_function()

                # Lift the block up
                move_function(pre_grasp_pose)

                # Get the target pose for placing the block
                target_point = self.get_actor_goal_pose(target_pose, target_pose_data)
                target_approach_direction = self.world_direction_dic['top_down']
                pre_place_pose = self.get_grasp_pose_from_goal_point_and_direction(block, block_data, endpose_tag=arm_tag, actor_functional_point_id=0, target_point=target_point, target_approach_direction=target_approach_direction, pre_dis=pre_dis)
                target_place_pose = self.get_grasp_pose_from_goal_point_and_direction(block, block_data, endpose_tag=arm_tag, actor_functional_point_id=0, target_point=target_point, target_approach_direction=target_approach_direction, pre_dis=0)

                # Move to the pre-place pose
                move_function(pre_place_pose)

                # Move to the place pose
                move_function(target_place_pose)

                # Open the gripper to place the block
                open_gripper_function()

                # Lift the arm up
                move_function(pre_place_pose)

            # Grasp and place block1
            grasp_and_place(block1, block1_data, block1_target_pose, block1_target_pose_data, pre_dis)

            # Avoid collision if necessary
            if self.get_actor_goal_pose(block1, block1_data)[0] > 0:
                avoid_collision_pose = self.get_avoid_collision_pose(avoid_collision_arm_tag='left')
                self.left_move_to_pose_with_screw(avoid_collision_pose)
            else:
                avoid_collision_pose = self.get_avoid_collision_pose(avoid_collision_arm_tag='right')
                self.right_move_to_pose_with_screw(avoid_collision_pose)

            # Grasp and place block2 on top of block1
            grasp_and_place(block2, block2_data, block1, block1_data, pre_dis)

            # Avoid collision if necessary
            if self.get_actor_goal_pose(block2, block2_data)[0] > 0:
                avoid_collision_pose = self.get_avoid_collision_pose(avoid_collision_arm_tag='left')
                self.left_move_to_pose_with_screw(avoid_collision_pose)
            else:
                avoid_collision_pose = self.get_avoid_collision_pose(avoid_collision_arm_tag='right')
                self.right_move_to_pose_with_screw(avoid_collision_pose)

            # Grasp and place block3 on top of block2
            grasp_and_place(block3, block3_data, block2, block2_data, pre_dis)

            # Avoid collision if necessary
            if self.get_actor_goal_pose(block3, block3_data)[0] > 0:
                avoid_collision_pose = self.get_avoid_collision_pose(avoid_collision_arm_tag='left')
                self.left_move_to_pose_with_screw(avoid_collision_pose)
            else:
                avoid_collision_pose = self.get_avoid_collision_pose(avoid_collision_arm_tag='right')
                self.right_move_to_pose_with_screw(avoid_collision_pose)
\end{pythoncode}

\end{document}